\pdfoutput=1

\documentclass[11pt]{article}

\usepackage{acl}

\usepackage{times}
\usepackage{latexsym}

\usepackage[T1]{fontenc}

\usepackage[utf8]{inputenc}

\usepackage{microtype}

\usepackage{inconsolata}

\title{A (More) Realistic Evaluation Setup for Generalisation of\\Community Models on Malicious Content Detection}

\author{
\bf
Ivo Verhoeven$^{\dagger}$, Pushkar Mishra$^{\ddagger}$, Rahel Beloch$^{\dagger}$\\
\bf
Helen Yannakoudakis$^{\mathsection}$, Ekaterina Shutova$^{\dagger}$\\
$\dagger$ ILLC, University of Amsterdam, The Netherlands\\
$\ddagger$ AI at Meta, London, United Kingdom\\
$\mathsection$ Dept. of Informatics, King's College London, United Kingdom\\
\normalsize{\texttt{\{name.lastname\}@uva.nl}, \texttt{pushkarmishra@meta.com}, \texttt{helen.yannakoudakis@kcl.ac.uk}}
}

\usepackage{layouts}
\usepackage{booktabs}
\usepackage{multicol}
\usepackage{multirow}
\usepackage{algorithm}
\usepackage{algpseudocode}
\usepackage{microtype, amsmath, amssymb}
\usepackage{booktabs}
\usepackage{array}
\usepackage{graphicx}
\usepackage{makecell}
\usepackage{adjustbox}
\usepackage{pdflscape}
\usepackage{rotating}
\usepackage{lipsum}
\usepackage{stfloats}

\usepackage{hyperref}

\hypersetup{
    pdfauthor={Ivo Verhoeven, Pushkar Mishra, Rahel Beloch, Helen Yannakoudakis, Ekaterina Shutova},
    pdftitle={A (More) Realistic Evaluation Setup for Generalisation of Community Models on Malicious Content Detection},
    pdfsubject={NAACL Submission},
}

\newcommand\numberthis{\addtocounter{equation}{1}\tag{\theequation}}

\begin{document}

\maketitle

\begin{abstract}
\textit{Community models} for malicious content detection, which take into account the context from a social graph alongside the content itself, have shown remarkable performance on benchmark datasets. Yet, misinformation and hate speech continue to propagate on social media networks. This mismatch can be partially attributed to the limitations of current evaluation setups that neglect the rapid evolution of online content and the underlying social graph. In this paper, we propose a novel evaluation setup for model generalisation based on our few-shot subgraph sampling approach. This setup tests for generalisation through few labelled examples in local explorations of a larger graph, emulating more realistic application settings. We show this to be a challenging \textit{inductive} setup, wherein strong performance on the training graph is not indicative of performance on unseen tasks, domains, or graph structures. Lastly, we show that graph meta-learners trained with our proposed few-shot subgraph sampling outperform standard community models in the inductive setup. We make our code publicly available.\footnote{Our anonymised code-base is available at: \url{https://github.com/rahelbeloch/meta-learning-gnns}}

\end{abstract}


\section{Introduction}
\label{sec:introduction}

The combination of connectivity and anonymity offered by social media inadvertently also provides the perfect channel for wide-spread dissemination of malicious content \cite{allcott_hunt_2017, mueller_flamesOfHate, eurobarometer, derronCommissionInformationDisorder2021}. By \textit{malicious content}, we consider any form of content detrimental to society, and focus on two specific forms: misinformation and hate speech. Mitigating the effect of malicious content requires content moderation, but this is a labour-intensive process that exacts an immense psychological toll on moderators \cite{Vosoughi2018, wiessnerJudgeOKs852021}. Consequently, automated detection of malicious content has seen increased academic interest \cite{ruffoStudyingFakeNews2023} and industry adoption.

\textit{Community models} for malicious content detection are models that operate on \textit{social graphs}, i.e., graphs of content and users. They 1) classify content nodes as malicious or not, 2) incorporate information from interacting users in the graph when doing so, and 3) leverage emergent network properties like homophily to boost detection performance \cite{maHomophilyNecessityGraph2021, hussainInterplayCommunitiesHomophily2021}. For community modelling on large-scale heterogeneous online communities, Graph Neural Networks (GNNs) are the architecture of choice \cite{phanFakeNewsDetection2023}. 

While community models for malicious content detection perform very well on benchmark datasets \cite{mishraAbusiveLanguageDetection2019, gongFakeNewsDetection2023}, social media platforms continue to grapple with such content. \citet{alharbiEvaluatingFakeNews2021} find that high accuracy in malicious content detection is not indicative of trustworthiness in general, as predictions often rely on dataset-specific features. Models also become outdated quickly as online content and communities evolve \cite{montiFakeNewsDetection2019}. \citet{bozarthBetterPerformanceEvaluation2020} find detection models to be brittle to changes in domain or publication date, a finding that \citet{nielsenMuMiNLargeScaleMultilingual2022} corroborate specifically for community models. Finally, \citet{phanFakeNewsDetection2023} conclude that ``we [have] no graph benchmark data for fake news detection in the graph learning community'' (p. 22), making any claims of state-of-the-art performance difficult to verify.

Evidently, there exists a mismatch in the performance of community models on research datasets and in more realistic application settings. Research datasets are static; they capture a view of the social graph weeks or months after relevant content has been introduced and spread. Current evaluation practices designed on static graphs are effectively \textit{transductive} \cite{graph_survey}, i.e., they implicitly assume that no new content or users will be introduced into the social graph, which leads to performance scores that obscure the discussed deficiencies.

In \textit{realistic} settings, new users and content nodes are constantly added to the social graph, and the topic or style of malicious content often radically changes. This is a property inherent to online content and communities \cite{adamicInformationEvolutionSocial2016, guoHowDoesTruth2021}. Thus, a successful community model should be able to rapidly adapt to domain shifts in content. Since labelling is prohibitively expensive, adaptation should occur in a few-shot manner. Furthermore, the community of interacting users also evolves. Initially, only a few users take note of some content, but as it gains traction, more and more users interact. To prevent harm from malicious content, detection must occur before wide-spread dissemination. This requires adaptation from a limited exploration of the social graph. Therefore, \textit{inductive} evaluation is needed.


In this paper, we seek to more realistically assess the generalisation capabilities of community models for malicious content detection, specifically making the following contributions:
\begin{enumerate}
    \vspace{-0.2em}
    \item We design a novel evaluation setup based on a few-shot subgraph sampling procedure that test inductive generalisation. The subgraphs are local, contain limited context, and have only a few labels.
    \vspace{-0.4em}
    \item We test a state-of-the-art community model under this novel evaluation setup on unseen graphs, domains, and tasks. We find them lacking the capacity to generalise.
    \vspace{-0.4em}
    \item We show that graph meta-learners trained with our few-shot sampling outperform standard community models in inductive evaluation.
\end{enumerate}

\section{Related Work}
\label{sec:related_work}

\subsection{Community Models}
\label{sec:related_gnns}
Community models have shown promise on static social graphs. Such models use social graphs to contextualise content by the users that interact with them, phrasing the detection task as node classification. \citet{mishraAuthorProfilingAbuse2018a, mishraAbusiveLanguageDetection2019} and \citet{montiFakeNewsDetection2019} find that GNNs over heterogeneous user--tweet graphs outperform models using only text or user-based features. \citet{chandraGraphbasedModelingOnline2020} show that relational GNNs---which directly model edge relations between different types of nodes---yield significant improvements over a range of baselines. \citet{shuNewsContentsRole2019} argue for the inclusion of publishers as another node type, with \citet{renAdversarialActiveLearning2021} also including topics. 

\citet{nguyenFANGLeveragingSocial2022} utilise temporal replies to model the dynamic user--content interactions. Temporal representations aid in early detection of malicious content \cite{jiancuiHeteroSCANSocialContext2021,songTemporallyEvolvingGraph2021}. However, they still assume static content. Others have focused on directly detecting actors posting the content \cite{botpercent, mehta-etal-2022-tackling}; we explicitly exclude actor modelling from our methodology since it mandates different ethical considerations \cite{mishra-etal-2021-modeling-users}.

\citet{gongFakeNewsDetection2023} provide a review of graph representation learning for malicious content detection. They also conclude that cross-domain generalisation remains an understudied problem for graph-based malicious content detection.

\subsection{Generalisable Content-only Models}
\label{sec:related_generalisation}
Developing malicious content detectors for cross-domain generalisation is receiving increased attention. For example, many task-aware domain adaptation approaches have been proposed \cite{zhangBDANNBERTBasedDomain2020, zhangLearningDetectFewShotFewClue2021, mosallanezhadDomainAdaptiveFake2022, linDetectRumorsMicroblog2022, yueContrastiveDomainAdaptation2022}. These methods are either ``aware'' of the distribution of representations in different datasets, or use external models to correct representations post-hoc. Generating representations that are invariant to domain shifts is a related direction \cite{dingMetaDetectorMetaEvent2022,huangMetapromptBasedLearning2023}.

Utilising (large) language models on unseen texts has also shown promising results \cite{leeFewshotFactCheckingPerplexity2021,chiuDetectingHateSpeech2022, alkhamissiReviewLanguageModels2022}. \citet{leeUnifyingMisinformationDetection2021} use multitask fine-tuning on RoBERTa \cite{liuRoBERTaRobustlyOptimized2019}, and show that few-shot adaptation on related, but unseen datasets improves performance over fine-tuning on individual tasks. \citet{yueMetaAdaptDomainAdaptive2023} specifically train content-only misinformation detectors to rapidly adapt. We, however, focus solely on community models for malicious content on social graphs, and to the best of our knowledge, are the first to do so.

\subsection{Subgraph Sampling \& Meta-learning}
\label{sec:related_metalearning}

Closely related to the idea of rapid adaptation to new tasks and domains is the field of meta-learning. Herein, models are trained to optimise themselves using a minimal amount of examples. In NLP, this has been investigated for document \cite{yuDiverseFewShotText2018,vanderheijdenMultilingualCrosslingualDocument2021}, sentence \cite{douInvestigatingMetaLearningAlgorithms2019, bansalLearningFewShotLearn2020}, and token-level tasks \cite{hollaLearningLearnDisambiguate2020}. For a thorough review, we refer the reader to \citet{leeMetaLearningIts2021,leeMetaLearningNatural2022}.

To operationalise meta-learning from subgraphs, \citet{huangGraphMetaLearning2021} propose G-Meta. They assume local subgraphs preserve information of a larger graph, such that training a GNN on relevant subgraphs can induce rapid adaptation from limited context. Other graph meta-learning procedures exist, however, they do not utilise episodes of local subgraphs. We refer the reader to \citet{mandalMetaLearningGraphNeural2021} and \citet{zhangFewShotLearningGraphs2022} for a comprehensive review of the field.

\begin{table*}[t]
    \centering
    \resizebox{0.9\textwidth}{!}{
\begin{tabular}{lccc}
    \toprule
    \footnotesize{\textbf{Dataset}} & \footnotesize{\textbf{GossipCop}} & \footnotesize{\textbf{CoAID}} & \footnotesize{\textbf{TwitterHateSpeech}} \\[-2pt]
    \midrule
    \footnotesize{Task} & \footnotesize{Rumour} & \footnotesize{Fake News} & \footnotesize{Hate Speech} \\[-2pt]
    \footnotesize{Domain} & \footnotesize{Celebrity Gossip} & \footnotesize{COVID-19} & \footnotesize{Entertainment} \\[-2pt]

    \makecell[l]{\footnotesize{Label} \\[-5pt] \footnotesize{Proportions}} &
        \begin{tabular}{c@{\hspace{0.75\tabcolsep}}c@{\hspace{0.75\tabcolsep}}}
             \footnotesize{True} & \footnotesize{Fake} \\[-5pt]
             \footnotesize{$77.12\%$} & \footnotesize{$22.88\%$}
        \end{tabular} &
        \begin{tabular}{c@{\hspace{0.75\tabcolsep}}c@{\hspace{0.75\tabcolsep}}}
             \footnotesize{True} & \footnotesize{Fake} \\[-5pt]
             \footnotesize{$94.72\%$} & \footnotesize{$5.28\%$}
        \end{tabular} & 
        \begin{tabular}{c@{\hspace{0.75\tabcolsep}}c@{\hspace{0.75\tabcolsep}}c@{\hspace{0.75\tabcolsep}}}
             \footnotesize{Racism} & \footnotesize{Sexism} & \footnotesize{None} \\[-5pt]
             \footnotesize{$11.97\%$} & \footnotesize{$19.43\%$} & \footnotesize{$68.60\%$}
        \end{tabular}
    \\[-2pt]
    \makecell[l]{\footnotesize{Doc--User} \\[-5pt] \footnotesize{Interaction}} & \footnotesize{Retweet} & \footnotesize{Retweet} & \footnotesize{Authorship} \\[-2pt]
    
    \midrule
    \footnotesize{\# Documents} & \footnotesize{17 617} & \footnotesize{947} & \footnotesize{16 201} \\[-2pt]
    \footnotesize{\# Users} & \footnotesize{29 229} & \footnotesize{4 059} & \footnotesize{1 875} \\[-2pt]
    \footnotesize{\# Edges} & \footnotesize{2 334 554} & \footnotesize{61 254} & \footnotesize{65 600} \\[-2pt]
    \bottomrule
\end{tabular}

    }
    \caption{Statistics of graph datasets post-processing. Further details are in Appendix \ref{appendix:dataset stats}.}
    \label{tab:graph_stats}
\end{table*}

\section{Datasets \& Tasks}
\label{sec:datasets}
We use three widely-adopted social graph datasets to train and evaluate community models. Social graph datasets are difficult to collect and degrade as users or content are moderated out. The first dataset, GossipCop, is used for pre-training. The other two datasets, CoAID and TwitterHateSpeech, are reserved for evaluating generalisation to unseen graphs. Table \ref{tab:graph_stats} provides some statistics, which are further complemented by Appendix \ref{appendix:dataset stats}. All datasets were rehydrated, i.e., rebuilt through the API, using the Twitter Academic API prior to May 2023. See \href{https://developer.twitter.com/en/developer-terms/more-on-restricted-use-cases}{`Redistribution of Twitter Content'}.

\vspace{1mm}
\noindent
\textbf{GossipCop} is one of two datasets introduced by \citet{shuFakeNewsNetDataRepository2019}. It comprises $20$k fact-checked celebrity rumour articles, and around $500$k interacting Twitter users. Labels correspond to the (now defunct) GossipCop fact-checking scores, covering a variety of (usually unreliable) publishers. Articles from a single trusted source, \href{https://www.eonline.com/}{E!Online}, were included to reduce class imbalance. Users are connected to articles and other users.

\vspace{1mm}
\noindent
\textbf{CoAID} contains articles from the first months of the COVID-19 pandemic, collected by \citet{cuiCoAIDCOVID19Healthcare2020}. We omit the ``social media'' category as most contain short, poorly formatted text. Fake news articles are labelled using a variety of fact-checking websites, whereas truthful news comes from (unverified) mainstream media outlets. After rehydration, this dataset is substantially smaller than when originally devised, with most lost documents corresponding to the fake class. Users are connected to articles and other users.

\vspace{1mm}
\noindent
\textbf{TwitterHateSpeech} differs in task, domain, and relational schema from the other datasets. Document nodes are tweets generated by Twitter users during a few seed events. \citet{waseemHatefulSymbolsHateful2016} identified prolific hate speech tweeters, and include their followers and followees in the graph. They manually labelled all tweets as racist, sexist or innocuous (none). Especially racist ($0.3$\%) users are over-represented, leading to few diverse regions in the graph. User--document connections indicate authorship (as opposed to tweet/retweet interactions). Users are also connected to other users.

\section{Methodology}
\label{sec:method}

A social graph, $\mathcal{G}=(\mathcal{V}, \mathcal{E})$, consists of a set of nodes $\mathcal{V}$ and a set of edges $\mathcal{E}$ indicating which nodes are incident to each other. A node's $r$-radius neighbourhood $\mathcal{N}_{r}(v)$ contains all other nodes that can be reached by paths of length $r$ (also called `hops', a series of incident nodes), and always includes $v$.

All datasets considered require modelling two node-types: documents ($\mathcal{V}^{\text{docs.}}$) and users ($\mathcal{V}^{\text{(users)}}$). Hence, the social graph $\mathcal{G}$ is heterogeneous. Document nodes $v$ contain exogenous features $x_{v}$ (i.e. the content representation), and have target labels $y_{v}\in\mathcal{Y}$. Documents are only connected to users ($\forall u\in\mathcal{N}_{1}(v), u\in \mathcal{V}^{\text{(users)}}$), whereas users may also be connected with other users based on their interactions or relations. Users are not labelled, and have no initial representation.

\begin{figure*}[t]
  \includegraphics[width=\textwidth, height=0.2\textheight]{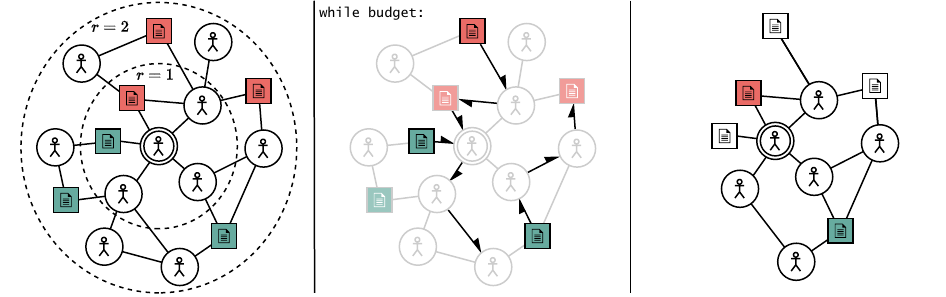}
  \caption{Support subgraph generation. Left: collect the $r$-radius neighbourhood of an anchor user. Middle: sub-sample using random walks from document nodes until reaching a maximum node count. Right: unmask document nodes inversely proportional to the number of subgraphs they appear in. Colours correspond to classes.}
  \label{fig:support_graph_sampling_procedure}
\end{figure*}

\subsection{Community Modelling}
\label{sec:graph_learning}


Community models for malicious content detection classify content nodes in a social graph, taking into account the graph context around them to make the prediction $f_{\theta}(x_{v};\mathcal{G})$. GNNs, a common representation learning framework, perform this contextualisation using non-linear message-passing schemes.

Some node $v$, at layer $l$, has as hidden state an aggregation of the representations in neighbouring nodes at layer $l-1$. We use Graph Attention Networks (GATs) \cite{velickovicGraphAttentionNetworks2018}, which employ an additive attention mechanism for the neighbourhood aggregation. Specifically:
\begin{equation}
  \label{eq:gat_layer}
  \mathbf{h}_{v}^{l}=\sigma({\textstyle \sum_{u\in\mathcal{N}_{1}(v)}}\alpha_{v,u}\mathbf{W}\mathbf{h}_{u}^{l-1})
\end{equation}
where $\sigma$ is a non-linear activation function. The attention weights $\alpha_{v,u}$ are computed as:
\begin{equation}
  \label{eq:gat_aggregation}
  \mathtt{softmax}(\sigma(\mathbf{a}^{\intercal}[\mathbf{W}\mathbf{h}_{v}^{l-1}\|\mathbf{W}\mathbf{h}_{u}^{l-1}]))
\end{equation}
where $\left[\cdot\|\cdot\right]$ is concatenation, while $\mathbf{a}$ and $\mathbf{W}$ are the projection matrices.

More expressive architectures than GATs exist and have been applied to malicious content detection \cite{chandraGraphbasedModelingOnline2020, gongFakeNewsDetection2023}. Such models, however, often introduce inductive biases specific to the task it seeks to solve. For example, relational attention aids performance but requires the relational schema to be consistent across datasets. Our evaluation and meta-learning setup is model agnostic.

Community modelling can be \textit{transductive} or \textit{inductive}. Transductive modelling assumes that the social graph remains static across training and prediction. Inductive modelling, instead, assumes the underlying social graph changes, in terms of content and users.  

Herein we differ from the definition common to graph learning applications. Usually, inductive graph learning 'only' assumes the nodes in the evaluation graph are unseen, with those nodes coming from the same underlying graph. As argued in the introduction, this does not apply to malicious content detection; the graph has shifted between training and deployment time. True inductive generalisation, therefore, requires generalisation to entirely different graphs. Currently, no graph datasets of malicious content exist that allow testing this manner of generalisation.

Due to the social network changing, inductive community models should not rely on superficial properties, like the content of malicious posts or specific user neighbourhoods, but rather leverage universal network properties. One such property is homophily: the tendency of nodes of a similar class to cluster together. We investigate the presence of homophily (or heterophily) in Appendix \ref{appendix:homophily}.



\subsection{Few-shot Subgraph Sampling}
\label{sec:ubs}
A successful community model for malicious content detection should be able to rapidly adapt to the constantly evolving social graph, even when presented with labelled examples. 

More formally, a community model, $f_{\theta}$, should be able to inductively learn to generalise from a limited exploration of a social graph $\mathcal{G}^{\mathcal{S}}\subset\mathcal{G}^{\prime}$ to make accurate predictions elsewhere $\mathcal{G}^{\mathcal{Q}}=\mathcal{G}^{\prime}-\mathcal{G}^{\mathcal{S}}$. In commonly-used meta-learning terminology, $\mathcal{S}$ would denote the support and $\mathcal{Q}$ the query set.

For malicious content detection specifically, the notion of `limited exploration' implies the following conditions for $\mathcal{G}^{\mathcal{S}}$:
\begin{enumerate}
    \vspace{-0.4em}
    \item \textbf{Locality}: all sampled document nodes come from the same graph region, due to a similar seed event, topic, or intended audience
    
    \vspace{-0.4em}
    \item \textbf{Limited Context}: moderation should precede wide-spread dissemination
    
    \vspace{-0.4em}
    \item \textbf{Few-shot}: labelling is expensive, therefore a minimal set of labels is available

    \vspace{-0.2em}
\end{enumerate}

Existing subgraph sampling procedures, like G-Meta, violate these conditions, especially `locality'. Labelled nodes are sampled independently, i.e., nodes can come from entirely different regions of the graph, which in our case, would imply entirely unrelated forms of content.

\begin{algorithm}[t]
      \caption{Few-shot Graph Sampling}\label{alg:user_based_graph_sampling}
      \begin{algorithmic}
            \Require Graph $\mathcal{G}=\left(\left(\mathcal{V}^{(\text{doc})}, \mathcal{V}^{(\text{user})}\right), \mathcal{E}\right)$
            \Require Central user $v$
            \Require Maximum support graph size, \texttt{budget} 
            \\

            \For{$r$ \texttt{in} \texttt{range}(2, 5)}
                  \State Get $r$-radius neighbourhood of $v$, $\mathcal{N}_{r}(v)$
                  \State Find nodes per class $\mathcal{V}^{\text{(doc)}}_{y}$
                  \If{$\mathtt{all}(|\mathcal{V}^{\text{(doc)}}_{y}|\geq k$-shot)}
                        \State \texttt{break}
                  \EndIf
            \EndFor
            \\

            \State Initialize empty support graph $\mathcal{G}^\prime$
            \While{$|\mathcal{G}^\prime|<$\texttt{budget}}
                  \State Pick class $y$ round-robin style
                  \State Sample a node $u$ from $\mathcal{V}^{\text{(doc)}}_{y}$
                  \State Generate random walk path, $\mathcal{G}^{\text{(path)}}$, from $u$
                  \State $\mathcal{G}^\prime\leftarrow \mathcal{G}^\prime\cup\mathcal{G}^{\text{(path)}}$
            \EndWhile
            \\ \\
            \Return $\mathcal{G}^\prime$ if it is a valid $k$-shot graph

      \end{algorithmic}
\end{algorithm}

To better conform to the listed desiderata, we perform user-centred sampling for generating $\mathcal{G}^{\mathcal{S}}$. Algorithm \ref{alg:user_based_graph_sampling} presents pseudocode for our sampling approach, which is also depicted graphically in Figure \ref{fig:support_graph_sampling_procedure}. Various graph statistics are provided in Appendix \ref{appendix:ubs}.

To ensure locality, we first sample an anchor user and collect the smallest $r$-hop neighbourhood that yields $k$ documents of each class. In Figure \ref{fig:support_graph_sampling_procedure}, the double-circled user represents the anchor. Then, to limit social context, we take random walks from the documents nodes into the subgraph. This process starts from a document node, and is repeated until a maximum number of nodes is reached. Bold arrows in the middle column of Figure \ref{fig:support_graph_sampling_procedure} show some random walks of length $3$.

For the training process, only $k$ document nodes of each class have their labels unmasked in a subgraph. Other document nodes are still allowed in the subgraph, but without labels. This is depicted in the right-most column of Figure \ref{fig:support_graph_sampling_procedure}.

Power-law distributed degrees of nodes is a property of social media networks. This means a few, very active users and their incident document nodes will be present in the majority of subgraphs. This reduces the diversity of support episodes and thus biases generalisability estimates. To reduce the effect of these users during the training process, document nodes are labelled inversely proportional to their frequency across all created subgraphs.

\subsection{Gradient-based Meta-learning}
\label{sec:gbml}
Community models learn a neighbourhood-aware mapping of a content node's input features to target labels. Community meta-learners, instead, use an initial set of weights to produce community models only after adaptation, i.e. learning a community model from several episodes of $\mathcal{G}^{\mathcal{S}}$ and $\mathcal{G}^{\mathcal{Q}}$. By using our few-shot subgraph sampling method to create episodes for meta-training, the community meta-learners are better suited to inductive generalisation.
 


We focus on a specific subclass of meta-learners, namely, gradient-based meta-learners. Model-Agnostic Meta-Learning (MAML), introduced by \citet{finn_ModelAgnosticMetaLearningFast2017}, is the most popular such learning framework. Its optimisation objective is:
\begin{equation}
    \label{eq:maml}
    \underset{\theta^{\text{(meta)}}}{\text{ min }} \mathbb{E}[\mathcal{L}(\mathbf{y}_{\text{Q}}, f_{\theta^{\text{(task)}}_{T_{\text{inner}}}}(\mathbf{x}_{\text{Q}};\mathcal{G}^{\mathcal{Q}}))]
\end{equation}

The induced update to $\theta^{\text{(meta)}}$ is called the \textit{outer-loop} update. The \textit{inner-loop} occurs during adaptation to the support set, using a pre-defined number of SGD updates, $t\in \{1, \ldots, T_{\text{inner}}\}$, with gradients
\begin{equation}\label{eq:maml_task_update}
    \nabla_{\theta^{\text{(task)}}_{t}}\mathcal{L}( \mathbf{y}_{\text{S}},f_{\theta^{\text{(task)}}_{t}}(\mathbf{x}_{\text{S}};\mathcal{G}^{\mathcal{S}}))
\end{equation}

This bi-level objective encourages the meta-learner's initial weights, $\theta^{\text{(meta)}}$, to learn to adapt to new tasks, $\theta^{\text{(task)}}$, using only $T_{\text{inner}}$ updates.


\paragraph{Prototypical Initialisation} MAML implicitly assumes a new permutation of classes in each episode and re-initialises the task-specific classification head during each outer-loop iteration. Prototypical Networks (ProtoNets) \cite{snellPrototypicalNetworksFewshot2017} are a non-gradient-based meta-learning alternative that does not utilise classification heads. Instead, support samples are used to form class prototypes $\mathbf{c}_{y}$: 
\begin{equation}
    \label{eq:prototype}
    \mathbf{c}_{y}=\dfrac{1}{k}{\textstyle \sum_{\{\mathbf{x}_{v}|y_{v}=y\}}} f_{\theta^{\text{(meta)}}}(\mathbf{x}_{\mathcal{S}};\mathcal{G}^{\mathcal{S}})
\end{equation}
which classify query samples based on their distance to the prototypes.
\begin{equation}
    \label{eq:prototypical_classification}
    p(y_{v}|\mathbf{x}_{v})=\text{softmax}(-d(f_{\theta^{\text{(meta)}}}(\mathbf{x}_{v}), \mathbf{c}_{y}))
\end{equation}

Per \citet{triantafillouMetaDatasetDatasetDatasets2020}, if using the Euclidean distance as $d$, this is equivalent to applying a linear projection $\mathbf{W}\mathbf{h}+\mathbf{b}$ with initialisation:
\begin{equation}
    \label{eq:protomaml_initialization}
    \mathbf{W}=2\mathbf{c}_{y},\quad\mathbf{b}=\|\mathbf{c}_{y}\|^2
\end{equation}

Using this reformulation, \citet{triantafillouMetaDatasetDatasetDatasets2020} propose ProtoMAML, an approach that extends MAML such that the classification head is now parameterised by Eq. \ref{eq:protomaml_initialization} and fully updatable during inner loop adaptation.

\subsection{Implementation Details}
Support graphs for episodes are sampled using the few-shot sampling procedure detailed in Sec. \ref{sec:ubs}. We use the lowest possible radius $r$ that satisfies the $k$-shot requirement. The maximum number of nodes in the support graph is $2048$. All classes provide an equal amount of root nodes of the random walk sub-sampling. For meta-training, $k=4$, and the query graph is generated by sampling a set of independent document nodes along with their $r=2$ neighbourhood. During (meta-)testing, all non-training nodes are used.

Document nodes use the time-pooled averaged token representations from the penultimate RoBERTa \cite{liuRoBERTaRobustlyOptimized2019} layer as initial representations. These are not trained further. Users nodes are initialised to all zeros, making them effectively anonymous and allowing for both transductive and inductive approaches. All models train end-to-end and do not include an auxiliary text-only classifier. Appendix \ref{appendix:hyperparameters} provides all additional modelling hyper-parameters.

Our GNN architecture is adapted from SAFER \cite{chandraGraphbasedModelingOnline2020}. It consists of $2$ ReLU activated GAT layers, each with $3$ attention heads. These are concatenated together and linearly projected before being fed into a $2$-layer MLP. We use dropout node-wise on the initial representations and element-wise on the layer representation and attention weights. We reduce the computational complexity of the GAT layers by merging successive projections in the attention layers \cite{brodyHowAttentiveAre2022}. The use of $2$ GAT layers means document nodes $v$ have as receptive field $\mathcal{N}_2(v)$. We optimise our models using AdamW \cite{kingmaAdamMethodStochastic2017, loshchilovDecoupledWeightDecay2019}.


In total, we experiment with models trained under $6$ different learning paradigms. The first two (\textsc{full} and \textsc{subgraphs}) are non-episodic baselines, trained transductively on the full graph or inductively on few-shot sampled subgraphs respectively. \textsc{full} mimics the current practice of training transductively without generalisation to unseen graphs in mind. \textsc{subgraphs} makes generalisation feasible and allows us to isolate the contribution of meta-learning.

The last four are graph meta-learners. We use two MAML variants, one with a classification head shared across episodes (\textsc{maml-lh}) and another where the classification head is randomly initialised at each episode (\textsc{maml-rh}). Appendix \ref{appendix:learned_and_reset_head} shows the effect of classifier head resetting on adaptation speed. We also train \textsc{protonet} and \textsc{protomaml} variants to evaluate the effect of prototypical initialisation on the classification head.

MAML-based outer-loop updates (Eq. \ref{eq:maml}) require computing second-order gradients, which is prohibitively expensive. Instead, we use a first-order approximation (foMAML \cite{finn_ModelAgnosticMetaLearningFast2017}) which usually does not significantly affect performance \cite{nicholFirstOrderMetaLearningAlgorithms2018,antoniouHowTrainYour2019}.

\begin{table*}[t]
    \centering
    \resizebox{\textwidth}{!}{

\begin{tabular}{clcccc}
    \toprule
    \multicolumn{2}{c}{\multirow{2}[2]{*}{\textbf{Method}}} & \multicolumn{2}{c}{\textbf{F1}} & \multirow{2}[2]{*}{\textbf{AUPR}} & \multirow{2}[2]{*}{\textbf{MCC}} \\
    \cmidrule(l{4pt}r{4pt}){3-4}
    \multicolumn{2}{c}{} & \textbf{Real} & \textbf{Fake} & & \\
    \midrule
    \multicolumn{2}{l}{SAFER \cite{chandraGraphbasedModelingOnline2020}}
          & & \small{\textbf{0.9453}} & &  \\[-2pt]
    \midrule
    \multicolumn{1}{c}{\multirow{2}[2]{*}{Baselines}} 
          & \textsc{text}  & \makecell{\small{0.8773}\\[-3pt] \footnotesize{(0.8628, 0.8918)}} & \makecell{\small{0.5963}\\[-3pt] \footnotesize{(0.5854, 0.6072)}} & \makecell{\small{0.6738}\\[-3pt] \footnotesize{(0.6570, 0.6905)}} & \makecell{\small{0.4767}\\[-3pt] \footnotesize{(0.4532, 0.5002)}} \\[-2pt]
          & \textsc{user id} & \makecell{\small{0.9423}\\[-3pt] \footnotesize{(0.9403, 0.9444)}} & \makecell{\small{0.7431}\\[-3pt] \footnotesize{(0.729, 0.7572)}} & \makecell{\small{0.8644}\\[-3pt] \footnotesize{(0.8556, 0.8733)}} & \makecell{\small{0.7164}\\[-3pt] \footnotesize{(0.7051, 0.7277)}} \\[-2pt]
    \midrule
    \multirow{6}[18]{*}{GAT}
          & \textsc{full}  & \makecell{\small{\underline{0.9672}}\\[-3pt] \footnotesize{(0.9615, 0.9728)}} & \makecell{\small{\underline{0.8920}}\\[-3pt] \footnotesize{(0.8754, 0.9086)}} & \makecell{\small{0.9450}\\[-3pt] \footnotesize{(0.9291, 0.9608)}} & \makecell{\small{\underline{0.8601}}\\[-3pt] \footnotesize{(0.8384, 0.8818)}} \\[-2pt]
          & \textsc{subgraphs} & \makecell{\small{0.9485}\\[-3pt] \footnotesize{(0.9406, 0.9563)}} & \makecell{\small{0.8496}\\[-3pt] \footnotesize{(0.8326, 0.8666)}} & \makecell{\small{\underline{0.9473}}\\[-3pt] \footnotesize{(0.9412, 0.9534)}} & \makecell{\small{0.8088}\\[-3pt] \footnotesize{(0.7886, 0.8289)}} \\[-2pt]
    \cmidrule{2-6}
          & \textsc{maml-lh}$^{\dag}$ & \makecell{\small{\textbf{0.9732}}\\[-3pt] \footnotesize{(0.9731, 0.9732)}} & \makecell{\small{\textbf{0.9092}}\\[-3pt] \footnotesize{(0.9091, 0.9094)}} & \makecell{\small{\textbf{0.9651}}\\[-3pt] \footnotesize{(0.9651, 0.9651)}} & \makecell{\small{\textbf{0.8828}}\\[-3pt] \footnotesize{(0.8826, 0.8831)}} \\[-2pt]
          & \textsc{maml-rh}$^{\dag}$ & \makecell{\small{0.8861}\\[-3pt] \footnotesize{(0.8776, 0.8946)}} & \makecell{\small{0.7559}\\[-3pt] \footnotesize{(0.7498, 0.7620)}} & \makecell{\small{0.9164}\\[-3pt] \footnotesize{(0.9136, 0.9192)}} & \makecell{\small{0.7108}\\[-3pt] \footnotesize{(0.7021, 0.7194)}} \\[-2pt]
          & \textsc{protonet}$^{\dag}$ & \makecell{\small{0.9205}\\[-3pt] \footnotesize{(0.9121, 0.9289)}} & \makecell{\small{0.8192}\\[-3pt] \footnotesize{(0.8116, 0.8268)}} & \makecell{\small{0.9175}\\[-3pt] \footnotesize{(0.9099, 0.9250)}} & \makecell{\small{0.7535}\\[-3pt] \footnotesize{(0.7384, 0.7686)}} \\[-2pt]
          & \textsc{protomaml}$^{\dag}$ & \makecell{\small{0.8921}\\[-3pt] \footnotesize{(0.8825, 0.9018)}} & \makecell{\small{0.7925}\\[-3pt] \footnotesize{(0.7857, 0.7994)}} & \makecell{\small{0.9255}\\[-3pt] \footnotesize{(0.922, 0.9290)}} & \makecell{\small{0.7263}\\[-3pt] \footnotesize{(0.7158, 0.7369)}} \\[-2pt]
    \bottomrule
\end{tabular}
    }
    \caption{Results on GossipCop. Brackets give the 90\% confidence interval. Bold values denote the best column score (where comparison is possible) and underlined the second-best. $\dag$ denotes $4$-shot episodic evaluation. SAFER results taken from \citet{chandraGraphbasedModelingOnline2020}; \textsc{full} is our re-implementation since they do not release their data splits.}
      \label{tab:gossipcop}
\end{table*}

\begin{figure*}[t]
  \includegraphics[width=\textwidth]{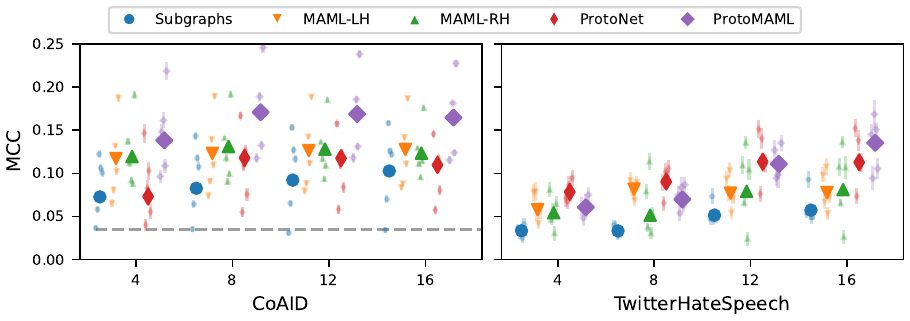}
  \caption{Generalisation of various models to CoAID and TwitterHateSpeech, in terms of MCC. See-through markers give the performance of each model instance, with error bars giving the 90\% CI. Solid markers give the performance averaged across model instances. Markers are offset to avoid overlap. The horizontal axis gives the support graph $k$-shot. The dashed gray line for CoAID gives the zero-shot performance of the Subgraphs model, i.e., direct domain transfer. Colours and shape both denote a model instance.}
  \label{fig:transfer_results}
\end{figure*}

\section{Experiments and Results}
\subsection{Experimental Setup}
\label{sec:experimental_setup}

We first assess within-dataset generalisation to unseen nodes using $5$-fold stratified cross-validation. The folds are strict, with no document appearing in more than one validation set. We only keep the largest connected component. User nodes can appear in multiple folds, but since they are all zero-initialised, they cannot influence nodes in other folds.

Models are trained on GossipCop and then evaluated on all three datasets. On GossipCop, we assess both the non-episodic and episodic models. On other datasets, we assess the episodic models only to ensure a few-shot generalisation setup. Within each episode, support nodes appear in the query graph but do not count towards classification performance metrics. This process is repeated $256$ times, with summary statistics computed for each of the $5$-fold model checkpoints. When aggregating over checkpoints, the inverse-variance weighted mean is used to estimate a common effect size (i.e., a fixed-effect meta-analysis \cite{schwarzerMetaAnalysis2015}).

To assess classification performance for each class in isolation, we use the F1-score. Matthews Correlation Coefficient (MCC) is used to assess holistically. Recent papers argue for the MCC as an informative metric, relatively robust to class imbalance \cite{chiccoAdvantagesMatthewsCorrelation2020,chiccoInvitationGreaterUse2022}. MCC values near 0 indicate random performance, values near 1 almost perfect performance, and negative values are worse than random. The Area Under the Precision-Recall curve (AUPR) is a multi-threshold metric, and can compare models on their ability to separate classes. We exclude it for CoAID and TwitterHateSpeech as there is no consistent way to do aggregate it in multi-class settings. Metrics are reported with 90\% confidence intervals.

All hyper-parameters used were tuned on GossipCop's validation sets. The tuning procedure, optimizer, meta-learning and evaluation hyper-parameters are described in  Appendix \ref{appendix:hyperparameters}.


\subsection{GossipCop Results}
\label{sec:gossipcop}
Here, we test generalisation on unseen nodes from the same graph. Beyond the non-episodic baselines already described, we have two additional baseline methods on GossipCop. The first, \textsc{text}, is a $2$ layer MLP on top of the initial document embeddings meant to test the added benefit of a graph inductive bias. The second, \textsc{user id}, classifies test documents based on neighbouring users' most linked document class in the train split. The already reasonable performance indicates high homophily.

The GAT-based models leverage both text and social features. \textsc{subgraphs} clearly performs worse than \textsc{full}. \textsc{maml-lh}, however, outperforms \textsc{full} even though it is inductive, demonstrating the generalisation power of meta-learning.

The lower three rows all include meta-learners which constantly re-initialise the classification head. Their performance is more in line with the non-episodic \textsc{subgraphs}, lagging considerably on fixed threshold metrics. This gap narrows in terms of AUPR, implying the final bias parameter may be to blame.

\subsection{Generalisation to Unseen Graphs}
\label{sec:transfer}

Figure \ref{fig:transfer_results} shows the performance of models ported to the other two datasets. 

Variance between the different model instances is large, although when aggregated their performance is relatively stable. \textsc{protomaml} proves to be the best model on both datasets, particularly at larger $k$-shot values. \textsc{protonet} shows competitive performance on TwitterHateSpeech (especially in lower $k$-shot settings), but is considerably worse on CoAID. Prototypical initialisation seems to aid generalisation, mitigating classification head learning. Regardless, meta-learning methods outperform the non-episodic \textsc{subgraphs} model on either dataset, indicating that training for rapid adaptation helps generalisation to new malicious content forms.

Transfer to CoAID from Gossipcop, is essentially a form of domain transfer. As such, we provide the zero-shot performance of the \textsc{subgraphs} model as a baseline value. Despite the similar task definition, adaptation is clearly required for generalisation, as evidenced by the near-random performance of \textsc{subgraphs} in the zero-shot setting, and the aggressive hyperparameters required (see Appendix \ref{appendix:coaid_transfer}.

Table \ref{tab:protomaml_f1} provides F1 scores for \textsc{protomaml} on each class. All-in-all, the highest achieved MCC was $0.1709$, for \textsc{protomaml} at $k=8$, corresponding to an F1-Fake of $0.1841$. While low relative to other F1 scores reported, this should be compared to a class prevalence of ~5\%.

Relative to random performance, the greatest negative outlier is TwitterHateSpeech's majority class, `None'. One possible explanation is the homophily pattern of TwitterHateSpeech (see Appendix \ref{appendix:homophily}). Whereas racist and sexist tweets are primarily homophilic in the query set, a large proportion of innocuous tweets are highly \textit{heterophilic}; i.e. these are contextualised by users predominantly authoring racist or sexist. The model is therefore more likely to err on those innocuous tweets, as their author shows a proclivity towards hate speech. Here, heterophily serves as noise. This is most likely an artefact of \citeauthor{waseemHatefulSymbolsHateful2016}'s data collection process, with prolific racist and sexist authors serving as the anchor around which the rest of the graph is built.

In general, the results here do not correlate with those found in Table \ref{tab:gossipcop}. Underperformers there show relatively better performance after adaptation to the other datasets. Hinting at overfitting to the GossipCop graph, this aligns with the line of argumentation presented in the introduction: performance on a single, static graph is not indicative of generalisation to emerging malicious content.


\subsection{Ablating GossipCop Pre-Training}
\label{sec:ablation}

To test the effect of GossipCop pre-training on generalisation to other datasets, we ablate \textsc{protomaml}'s pre-trained weights, and repeat the evaluation under re-initialised weights. A comparison in terms of MCC is provided in Table \ref{tab:protomaml_mcc}. On CoAID, \textsc{protomaml} outperforms \textsc{protomaml-reset} at all $k$-shot values. On TwitterHateSpeech, this only happens at the larger $k$-shot values, with \textsc{protomaml}'s MCC performance increase outpacing its \textsc{reset} counterpart. 

While low, comparing the performance on each class individually (Tables \ref{tab:protomaml_f1} and \ref{tab:twitter_random_transfer}), \textsc{protomaml} is able to increase its performance on all classes simultaneously, whereas \textsc{protomaml-reset} only does so for the racist class, degrading performance on sexist and innocuous documents.

Regardless, the fact that a model trained specifically with generalisation in mind is barely able to outperform one with random weights is striking, and speaks to the inadequacy of GossipCop as an evaluation dataset. Furthermore, near perfect performance on unseen nodes of the pre-training graph does not imply inductive generalisation to new graphs.

\begin{table}[t]
  \centering
\begin{tabular}{ccc@{\hspace{3.5\tabcolsep}}cccccccc}
    \toprule
    \multirow{2}[2]{*}{\textbf{k}} & \multicolumn{2}{c}{\textbf{CoAID}~~~~~~} & \multicolumn{3}{c}{\textbf{TwitterHateSpeech}} \\
    \cmidrule(l{4pt}r{20pt}){2-3}\cmidrule(r{4pt}){4-6}
          & Real  & Fake  & Racist & Sexist & None \\
    \midrule
    4     & \small{0.7734} & \small{0.1762} & \small{0.1763} & \small{0.2181} & \small{0.3585} \\
    8     & \small{0.8245} & \small{0.1841} & \small{0.1934} & \small{0.2148} & \small{0.3530} \\
    12    & \small{0.8245} & \small{0.1732} & \small{0.2545} & \small{0.2554} & \small{0.3503} \\
    16    & \small{0.8321} & \small{0.1599} & \small{0.3021} & \small{0.3077} & \small{0.3163} \\
    \midrule
    B & \small{0.6545} & \small{0.0955} & \small{0.1932} & \small{0.2799} & \small{0.5784} \\
    \bottomrule
\end{tabular}
    \caption{F1 scores achieved by \textsc{protomaml} during generalisation to the auxiliary datasets. Row B provides F1 scores for a random classifier \cite{flachPrecisionRecallGainCurvesPR2015}. This table is complemented by Appendix \ref{appendix:extended_transfer}.
    }
    \label{tab:protomaml_f1}
\end{table}

\section{Conclusion}
\label{sec:conclusion}

This paper proposes a more realistic evaluation setup for community models on malicious content detection. We highlight several properties of evolving social graphs that are especially neglected: expensive labelling, limited context, and emerging content and users. Experiments verified our motivation, with performance on a single, static dataset in a transductive manner bearing little resemblance to performance during few-shot inductive generalisation. 

Our proposed few-shot subgraph sampling approach presented in Section \ref{sec:ubs} is tailored to social media graphs and allows generalisation of community models to new networks, domains, and tasks. While standard community models performed poorly, incorporating our sampling procedure in graph meta-learners aided generalisation. Particularly promising are models with prototypical initialisation. 


Ultimately, our results suggest that malicious content detection using community models is not `solved', despite some models achieving near perfect evaluation scores. Current evaluation procedures neglect critical properties of malicious content, and models tested under these conditions will not prove useful in realistic deployment settings. This is a regrettable consequence, considering the high-stakes nature of malicious content detection. Much like the trend occurring in the content-only malicious content detection literature (see Section \ref{sec:related_generalisation}), we hope this work will lead to similar follow-up work for community models.

An open problem, warranting further investigation, is the application of meta-learning to class imbalanced datasets. Common to malicious content detection, class imbalance imposes a severe penalty on meta-learners that reset their classification heads.

\begin{table}[t]
  \centering
\begin{tabular}{ccc@{\hspace{3.5\tabcolsep}}cc}
    \toprule
        \multirow{2}[2]{*}{\textbf{k}} & \multicolumn{2}{c}{\textbf{CoAID}~~~~~~} & \multicolumn{2}{c}{\textbf{TwitterHateSpeech}} \\
        \cmidrule(l{4pt}r{20pt}){2-3}\cmidrule(r{4pt}){4-5}
              & Trained & Reset & Trained & Reset \\
    \midrule
        4     & \small{0.1383} & \small{0.1191} & \small{0.0607} & \small{0.0767} \\
        8     & \small{0.1709} & \small{0.1398} & \small{0.0699} & \small{0.0868} \\
        12    & \small{0.1689} & \small{0.1304} & \small{0.1109} & \small{0.1025} \\
        16    & \small{0.1646} & \small{0.1212} & \small{0.1354} & \small{0.1052} \\
    \midrule
    B & \small{0.0000} & \small{0.0000} & \small{0.0000} & \small{0.0000} \\
    \bottomrule
\end{tabular}
    \caption{MCC scores achieved by \textsc{protomaml} (Trained) and \textsc{protomaml-reset} (Reset) during generalisation to the auxiliary datasets. This table is complemented by Appendix \ref{appendix:random_transfer}.}
    \label{tab:protomaml_mcc}
\end{table}

\section{Limitations}
\label{sec:limitations}

While we took care to increase the diversity of the training data (user-centred sampling, distributing the labels, adding dropout throughout models), ultimately, the diversity is limited by the underlying graph dataset. GossipCop is large, but it contains only a single task and a relatively uniform structure. Ideally, multiple, distinct graph datasets would be used in meta-training. However, few such datasets are available. For meta-learning, task diversity might be a critical factor in `learning-to-learn', analogous to data diversity being critical in standard machine-learning setups. As such, our meta-learners are likely operating below capacity.

\citet{phanFakeNewsDetection2023} made the conclusion that no common benchmarks for community models of malicious content are currently in use. This work, despite making a step towards more realistic evaluation of such models, does not improve this situation. The presented performance metrics are in line with related work, but meaningful comparison will only be possible with the publication and adoption of open-access benchmark datasets. This seems an unrealistic short-term aspiration at the time of writing.

\section{Ethical Considerations}
\label{sec:ethics}
Following the work of \citet{mishra-etal-2021-modeling-users}, we take some steps to ensure that our experimental setup addresses the ethical considerations that may arise when modelling users and communities. The authors highlight the following three considerations that apply to our work:
\begin{itemize}
    \item \textit{Personal vs. Population-level trends}, i.e., are generalisations being made from personal traits to population-level trends
    \item \textit{Bias in datasets}, i.e., is there demographic, comment distribution, or label bias in the dataset(s) being used?
    \item \textit{Purpose}, i.e., is the purpose of the modelling being done to classify content as malicious or users and communities too? 
\end{itemize}

In order to tackle the \textit{comment distribution} bias whereby the majority of documents may belong to a small number of users, we remove users with more than a certain number of documents from the dataset (where possible). Furthermore, to counter the \textit{label distribution} bias where we only pick documents of a particular class from a specific user, we do user-centred sampling, incorporating the entire neighbourhood of a user in the user-document graph. We initialise all users to the same zero-embedding, ensuring that we do not generalise personal traits to population-level trends. Lastly, in our work, we solely leverage the user-document graph to be able to better classify the documents, not the users themselves as malicious, hence having a clear purpose to advance malicious content detection.

\bibliography{anthology, zotero}

\clearpage
\appendix
\begin{table*}[t]
\centering
\begin{tabular}{m{1.5in}|m{1.5in}|m{1.5in}|m{1.5in}}
     \toprule
     \multicolumn{1}{c}{\textbf{~~~~~Credibility~~~~~}} & \multicolumn{1}{c}{\textbf{2017}} & \multicolumn{1}{c}{\textbf{2020}} & \multicolumn{1}{c}{\textbf{2022}} \\
    \midrule
     \multicolumn{1}{c|}{High} 
        & {\footnotesize Trump says he'll allow Kennedy assassination files to be released} 
        & {\footnotesize Pelosi: Proposal on COVID-19 relief is ``one step forward, two steps back''}
        & {\footnotesize MPs insist fans heading to World Cup must not be priced out of enjoying a beer }
        \\
     \midrule
     \multicolumn{1}{c|}{Low}
        & {\footnotesize Russian Email Uncovered… Reveals What Really Happened at Trump Jr and Russia Mtg}
        & {\footnotesize Pence Destroys Biden's Record: He'd Have Killed 2 Million People Fighting COVID}
        & {\footnotesize Joe Biden: 'Inflation Is Going to Get Worse' if Republicans Win Despite Core Inflation Rising to 40-year High}
        \\
    \bottomrule
\end{tabular}
\caption{A series of headlines taken from the NELA-GT corpora. All articles come from around Oct. 15th in their respective year.}
\label{tab:nela_example}

\vspace{1em}

\begin{tabular}{m{1in}|m{1.5in}|m{1.5in}|m{1.5in}}
     \toprule
     \multicolumn{1}{c}{\textbf{Rumour Status}} & \multicolumn{1}{c}{\textbf{Anchor}} & \multicolumn{1}{c}{\textbf{Near}} & \multicolumn{1}{c}{\textbf{Far}} \\
     \midrule
     \multicolumn{1}{c|}{True}
        & {\footnotesize It's been 10 years since Heath Ledger died of an accidental drug overdose. Since}
        & {\footnotesize Who cut the head off of the General Pickens statue and what is going on in the Cooper}
        & {\footnotesize When Pretty Little Liars and Teen Wolf collide we get Truth or Dare. Well, not really,}
        \\
    \midrule
     \multicolumn{1}{c|}{Fake} 
        & {\footnotesize A man reportedly got his finger bitten off at a Beyoncé concert! The shocking twist: It wasn't} 
        & {\footnotesize A white witch from North London has urged Hollywood star Angelina Jolie to cease}
        & {\footnotesize Is Cher concerned Chaz Bono will die from his weight issues? That’s the claim from}
        \\
    \bottomrule
\end{tabular}
\caption{A similar set of snippets taken from the GossipCop dataset. The left column provides an anchor document, the middle another document near the anchor, and the right column a document far away from the anchor.}
\label{tab:gossipcop_example}
\end{table*}

\section{Motivating Examples}
\textit{DISCLAIMER: the chosen examples were taken verbatim from various malicious content corpora. They do not reflect the views of the authors.}

As established in Section \ref{sec:introduction}, malicious content and its social context evolves. This can happen quickly, and results in text that is very different from already seen forms of malicious content.

Table \ref{tab:nela_example} shows such change, depicting titles from articles published by low (i.e. those that often publish severely biased or false news) and high credibility news sources, as found in the NELA-GT corpora \cite{DVN/ZCXSKG_2019, DVN/CHMUYZ_2021, DVN/AMCV2H_2023}.

Current models are adept at filtering out malicious content as in their training datasets, but quickly degrade when presented with novel content. For example, Table \ref{tab:nela_example} shows substantial high-level semantic change across the years. Models relying on surface-level features will fail as new events spawn new content.

Currently, no existing social-network malicious content dataset captures this level of evolution. In fact, existing datasets are completely static, implicitly assuming the full network (users, content, and their connections) will be available at inference time. This paper argues that this assumption is a significant reason why malicious content continues to propagate unabated, despite the impressive classification scores reported in earlier work.

To showcase this lack of diversity, Table \ref{tab:gossipcop_example} depicts a similar array of texts, sampled from the GossipCop graph. While a variety of topics are discussed, the overarching subject remains the same. Comparatively, relying on surface level features can already lead to strong classification performance. Simply put, in this dataset, models need not account for evolving content.

In lieu of large, temporally diverse graph datasets, we propose an evaluation framework that approximates these effects through generalisation to new graphs. Requiring adaptation from a minimal set of examples, with limited social context will serve as a much better measure of the inference time performance of community models for malicious content detection. There is little point in good within dataset performance, when unseen content forms are free to cause harm.

Put otherwise, when it comes to malicious content detection, we want models that are able to filter out tomorrow's hate speech posts and fake news articles, not those seen yesterday.

\section{Additional Details on Datasets Used}
\label{appendix:dataset stats}
Rehydrating many years after the datasets were released, not all documents and users could be recovered. This results in empty, missing or isolated documents. The collected graph datasets are thus subgraphs of the one presented in \citet{shuFakeNewsNetDataRepository2019}, \citet{cuiCoAIDCOVID19Healthcare2020} and \citet{waseemHatefulSymbolsHateful2016}. The additional preprocessing steps needed are described here:
\begin{enumerate}
      \item \textbf{Tokenization}: all found documents were collected and tokenized. Any empty documents, or documents yielding only special tokens, were removed
      
      \item \textbf{Document-User Interactions}: where possible, user and doc-user interactions were collected. One issue with GossipCop, as identified by \citeauthor{shuFakeNewsNetDataRepository2019}, is the inclusion of bots. These `users' tend to disproportionately interact with documents of a single class, both in terms of volume and proportion. Therefore, following the recommendation made by \citet{chandraGraphbasedModelingOnline2020}, users sharing more than 30\% of the documents of any class were removed. The type of doc-user interactions in TwitterHateSpeech differs, resulting in a very small pool of racist users, so this restriction was relaxed. Documents without \textit{any} user interactions, were also removed.
      
      \item \textbf{User-user Interactions}: all remaining users and their interaction with other users were parsed at this point. To further reduce the number of bots, the top 1\% most active users were removed on GossipCop. Then, again only on GossipCop, to further sparsify the graph, only the top 30k users were kept. Isolated documents were once again removed.
      
\end{enumerate}

The effect of each filtering step and additional statistics of the dataset graph prior to generating episodic subgraphs, are presented in Table \ref{tab:appendix_dataset_stats}.

\begin{table*}[p]
    \centering
    \begin{tabular}{clccc}
        \toprule
        \multicolumn{1}{c}{\textbf{Group}} & \multicolumn{1}{c}{\textbf{Metric}} & \multicolumn{1}{c}{\textbf{GossipCop}} & \multicolumn{1}{c}{\textbf{CoAID}} & \multicolumn{1}{c}{\textbf{Twitter Hate Speech}} \\
        \midrule
        \multicolumn{1}{c}{\multirow{4}[0]{*}{\textbf{Typology}}} & Task & \makecell{\small{Rumour} \\[-3pt] \small{Verification}} & \makecell{\small{Fake News} \\[-3pt] \small{Detection}} & \makecell{\small{Hate Speech} \\[-3pt] \small{Classification}} \\[-2pt]
        & Domain & \small{Celebrity Gossip} & \small{COVID-19} & \small{Entertainment} \\[-2pt]
        & Labels & 
            \begin{tabular}{c@{\hspace{0.25\tabcolsep}}c@{\hspace{0.25\tabcolsep}}}
                 \small{True} & \small{Fake} \\[-5pt]
                 \scriptsize{77.12\%} & \scriptsize{22.88\%}
            \end{tabular} &
            \begin{tabular}{c@{\hspace{0.25\tabcolsep}}c@{\hspace{0.25\tabcolsep}}}
                 \small{True} & \small{Fake} \\[-5pt]
                 \scriptsize{94.72\%} & \scriptsize{5.28\%}
            \end{tabular} & 
            \begin{tabular}{c@{\hspace{0.25\tabcolsep}}c@{\hspace{0.25\tabcolsep}}c@{\hspace{0.25\tabcolsep}}}
                 \small{Racism} & \small{Sexism} & \small{None} \\[-5pt]
                 \scriptsize{11.97\%} & \scriptsize{19.43\%} & \scriptsize{68.60\%}
            \end{tabular}
        \\[-2pt]
        & Doc--user Interactions & \small{Retweet} & \small{Retweet} & \small{Authorship} \\[-2pt]
        \midrule
            \multicolumn{1}{c}{\multirow{3}[0]{*}{\textbf{Length}}} & Mean  & \footnotesize{352.99} & \footnotesize{71.34} & \footnotesize{24.42} \\
            & Std. Dev & \footnotesize{165.4} & \footnotesize{42.44} & \footnotesize{9.38} \\
            & Median   & \footnotesize{405} & \footnotesize{93} & \footnotesize{25} \\
        \midrule
        \multicolumn{1}{c}{\multirow{2}[0]{*}{\textbf{Missing Documents}}} & Not Found & \footnotesize{1168} & \footnotesize{0} & \footnotesize{0} \\
            & Empty & \footnotesize{488} & \footnotesize{0} & \footnotesize{0} \\
        \midrule
        \multicolumn{1}{c}{\multirow{5}[-2]{*}{\textbf{Users}}}
            & \#Users (pre filter) & \footnotesize{549225} & \footnotesize{5524} & \footnotesize{1875} \\
            & Unique Users - 0 & \footnotesize{215072} & \footnotesize{5062} & \footnotesize{5} \\
            & Unique Users - 1 & \footnotesize{384760} & \footnotesize{462} & \footnotesize{527} \\
            & Unique Users - 2 & \footnotesize{N/A} & \footnotesize{N/A} & \footnotesize{1648} \\
            & Too active & \footnotesize{36} & \footnotesize{0} & \footnotesize{0} \\
        \midrule
        \multicolumn{1}{c}{\multirow{8}[0]{*}{\textbf{User-Doc Interaction}}} 
        & Isolated Docs-0 & \footnotesize{1261} & \footnotesize{3635} & \footnotesize{0} \\
        & Isolated Docs-1 & \footnotesize{224} & \footnotesize{875} & \footnotesize{0} \\
        & Isolated Docs-2 & N/A   & N/A   & \footnotesize{0} \\
        & Mean  & \footnotesize{2.3} & \footnotesize{1.06} & \footnotesize{8.64} \\
        & Std. Dev & \footnotesize{31.27} & \footnotesize{0.41} & \footnotesize{147.49} \\
        & Median   & \footnotesize{1} & \footnotesize{1} & \footnotesize{1} \\
        & $\mathbb{E}[\log(x)]$ & \footnotesize{0.25} & \footnotesize{0.03} & \footnotesize{0.49} \\
        & Geom. Mean & \footnotesize{1.29} & \footnotesize{1.04} & \footnotesize{1.63} \\
        \midrule
        \multicolumn{1}{c}{\multirow{4}[0]{*}{\textbf{User Truncation}}} & Most Active & \footnotesize{5213} & \footnotesize{0} & \footnotesize{0} \\
        & Least Active & \footnotesize{486087} & \footnotesize{0} & \footnotesize{0} \\
        & \# Doc. Incident & \footnotesize{27148} & \footnotesize{4284} & \footnotesize{1875} \\
        & \# Doc. Non-incident & \footnotesize{2081} & \footnotesize{0} & \footnotesize{0} \\
        \midrule
        \multicolumn{1}{c}{\multirow{5}[0]{*}{\textbf{User Degrees}}} & Mean  & \footnotesize{3445.78} & \footnotesize{2313.63} & \footnotesize{16.71} \\
        & Std. Dev & \footnotesize{1553.34} & \footnotesize{2373.26} & \footnotesize{149.15} \\
            & Median   & \footnotesize{3199} & \footnotesize{1281} & \footnotesize{4} \\
            & $\mathbb{E}[\log(x)]$ & \footnotesize{8.03} & \footnotesize{6.95} & \footnotesize{1.54} \\
            & Geom. Mean & \footnotesize{3070.33} & \footnotesize{1042.83} & \footnotesize{4.65} \\
        \midrule
        \multicolumn{1}{c}{\multirow{5}[0]{*}{\textbf{Graph}}}
            & \#Nodes & \footnotesize{46846} & \footnotesize{5006} & \footnotesize{18076} \\
            & User--doc Edges & \footnotesize{284757} & \footnotesize{4520} & \footnotesize{16201} \\
            & User--user Edges & \footnotesize{859097} & \footnotesize{23604} & \footnotesize{7561} \\
            & Total Edges (uni) & \footnotesize{2334554} & \footnotesize{61254} & \footnotesize{65600} \\
            & Density & \footnotesize{2.13E-03} & \footnotesize{4.89E-03} & \footnotesize{4.00E-04} \\
        \bottomrule
    \end{tabular}
    \caption{Dataset filtering and additional statistics on documents and interactions.}
    \label{tab:appendix_dataset_stats}
\end{table*}

\section{Additional Information on Few-shot Subgraph Sampling}
\label{appendix:ubs}

Algorithm \ref{alg:user_based_graph_sampling} presents the few-shot subgraph sampling pseudocode. Although the models used can only aggregate information from at most an $r=2$ radius subgraph, the initial graph can be expanded to larger $r$. This is to ensure at least $k$-shot examples of each label is present.  For the used graph $r=5$ usually contains the vast majority of the graph, and would only be needed for extremely sparse areas. In practise, however, the $k$-shot was achieved by $r=3$ in all situations.

The random walk subsampling dramatically reduces the number of nodes and edges present in the subgraph. A similar strategy was employed by GraphSAINT \cite{zengGraphSAINTGraphSampling2020}, resulting in efficiency improvements for a variety of inductive graph learners. We set our random walk length to 5 for all experiments, preferring fewer document nodes (with smaller walk length requiring more roots to get to the same node budget). An approximate budget of 2048 was used during training and evaluation. 

Important statistic on the produced subgraphs, for both the support and query sets, are presented in Tables \ref{tab:ubs_support_stats} and \ref{tab:ubs_query_stats}.

\begin{table*}[p]
    \centering
    \begin{tabular}{clccccc}
        \toprule
        \multirow{2}[2]{*}{\textbf{Dataset}} & \multicolumn{1}{c}{\multirow{2}[2]{*}{\textbf{Metric}}} & \multicolumn{5}{c}{\textbf{Support}} \\
        \cmidrule(l{4pt}r{4pt}){3-7}
            &       & Mean  & Std. Dev. & Q25   & Median & Q75 \\
        \midrule
        \multirow{7}[-2]{*}{\begin{sideways}GossipCop\end{sideways}}
        & \small{\#Nodes} & \footnotesize{1994.05} & \footnotesize{242.98} & \footnotesize{2049} & \footnotesize{2050} & \footnotesize{2051} \\[-2pt]
        & \small{\#Edges} & \footnotesize{111293.61} & \footnotesize{41409.85} & \footnotesize{84358.75} & \footnotesize{115236} & \footnotesize{130180.25} \\[-2pt]
        & \small{\#Docs.} & \footnotesize{200.62} & \footnotesize{232.94} & \footnotesize{40.25} & \footnotesize{92} & \footnotesize{256} \\[-2pt]
        & \small{Density} & \footnotesize{0.06} & \footnotesize{0.02} & \footnotesize{0.04} & \footnotesize{0.06} & \footnotesize{0.06} \\[-2pt]
        & \small{Prop. Docs.} & \footnotesize{9.90\%} & \footnotesize{11.35\%} & \footnotesize{2.15\%} & \footnotesize{4.59\%} & \footnotesize{12.58\%} \\[-2pt]
        & \small{Deg. Cent.} & \footnotesize{1.08E-02} & \footnotesize{6.53E-03} & \footnotesize{7.39E-03} & \footnotesize{9.51E-03} & \footnotesize{1.25E-02} \\[-2pt]
        & \small{Eigen Cent.} & \footnotesize{2.40E-03} & \footnotesize{2.44E-03} & \footnotesize{1.65E-03} & \footnotesize{2.10E-03} & \footnotesize{2.75E-03} \\[-2pt]
        \midrule
        \multirow{7}[-2]{*}{\begin{sideways}CoAID\end{sideways}}
        & \small{\#Nodes} & \footnotesize{1796.66} & \footnotesize{448.43} & \footnotesize{1765.25} & \footnotesize{2048} & \footnotesize{2049} \\[-2pt]
        & \small{\#Edges} & \footnotesize{30464.29} & \footnotesize{6866.59} & \footnotesize{27552} & \footnotesize{33672.5} & \footnotesize{35229.25} \\[-2pt]
        & \small{\#Docs.} & \footnotesize{208.55} & \footnotesize{107.16} & \footnotesize{114} & \footnotesize{243} & \footnotesize{302} \\[-2pt]
        & \small{Density} & \footnotesize{0.02} & \footnotesize{0.03} & \footnotesize{0.02} & \footnotesize{0.02} & \footnotesize{0.02} \\[-2pt]
        & \small{Prop. Docs.} & \footnotesize{10.88\%} & \footnotesize{4.62\%} & \footnotesize{7.25\%} & \footnotesize{12.11\%} & \footnotesize{14.80\%} \\[-2pt]
        & \small{Deg. Cent.} & \footnotesize{7.46E-03} & \footnotesize{9.75E-03} & \footnotesize{3.42E-03} & \footnotesize{4.01E-03} & \footnotesize{6.34E-03} \\[-2pt]
        & \small{Eigen Cent.} & \footnotesize{3.18E-03} & \footnotesize{3.88E-03} & \footnotesize{1.09E-03} & \footnotesize{1.86E-03} & \footnotesize{3.18E-03} \\[-2pt]
        \midrule
        \multirow{7}[-2]{*}{\begin{sideways}TwitterHS\end{sideways}}
        & \small{\#Nodes} & \footnotesize{2050.55} & \footnotesize{2.45} & \footnotesize{2050} & \footnotesize{2051} & \footnotesize{2051} \\[-2pt]
        & \small{\#Edges} & \footnotesize{10381.76} & \footnotesize{515.17} & \footnotesize{10070.25} & \footnotesize{10312} & \footnotesize{10564.25} \\[-2pt]
        & \small{\#Docs.} & \footnotesize{1716.29} & \footnotesize{27.05} & \footnotesize{1707.25} & \footnotesize{1721} & \footnotesize{1732} \\[-2pt]
        & \small{Density} & \footnotesize{0} & \footnotesize{0} & \footnotesize{0} & \footnotesize{0} & \footnotesize{0.01} \\[-2pt]
        & \small{Prop. Docs.} & \footnotesize{83.70\%} & \footnotesize{1.33\%} & \footnotesize{83.26\%} & \footnotesize{83.91\%} & \footnotesize{84.45\%} \\[-2pt]
        & \small{Deg. Cent.} & \footnotesize{1.46E-03} & \footnotesize{1.88E-06} & \footnotesize{1.46E-03} & \footnotesize{1.46E-03} & \footnotesize{1.46E-03} \\[-2pt]
        & \small{Eigen Cent.} & \footnotesize{2.48E-04} & \footnotesize{1.20E-03} & \footnotesize{1.40E-04} & \footnotesize{1.50E-04} & \footnotesize{1.62E-04} \\[-2pt]

        \bottomrule
    \end{tabular}%
    \caption{Additional statistics on the subgraphs generated by the proposed sampling procedure on the support graphs. Each row header gives the dataset. The metrics provided include: the number of nodes, number of edges, number of document nodes, the graph density, the proportion of document nodes, the degree centrality of document nodes, and the eigen centrality of document nodes. \\[1em]}
    \label{tab:ubs_support_stats}
    
    \begin{tabular}{clccccc}
        \toprule
        \multirow{2}[2]{*}{\textbf{Dataset}} & \multicolumn{1}{c}{\multirow{2}[2]{*}{\textbf{Metric}}} & \multicolumn{5}{c}{\textbf{Query}} \\[-2pt]
        \cmidrule(l{4pt}r{4pt}){3-7}  
            &       & Mean  & Std. Dev. & Q25   & Median & Q75 \\
        \midrule
        \multirow{7}[-2]{*}{\begin{sideways}GossipCop\end{sideways}}
            & \small{\#Nodes} & \footnotesize{3046.03} & \footnotesize{2821.88} & \footnotesize{609.75} & \footnotesize{2697.5} & \footnotesize{4779.25} \\[-2pt]
            & \small{\#Edges} & \footnotesize{75547.65} & \footnotesize{187261.29} & \footnotesize{5549} & \footnotesize{18839} & \footnotesize{48589} \\[-2pt]
            & \small{\#Docs.} & \footnotesize{2181.1} & \footnotesize{1976.14} & \footnotesize{281} & \footnotesize{1616} & \footnotesize{4108.5} \\[-2pt]
            & \small{Density} & \footnotesize{0.04} & \footnotesize{0.09} & \footnotesize{0} & \footnotesize{0.01} & \footnotesize{0.04} \\[-2pt]
            & \small{Prop. Docs.} & \footnotesize{68.76\%} & \footnotesize{31.75\%} & \footnotesize{43.82\%} & \footnotesize{85.18\%} & \footnotesize{94.75\%} \\[-2pt]
            & \small{Deg. Cent.} & \footnotesize{4.32E-04} & \footnotesize{8.01E-04} & \footnotesize{1.32E-04} & \footnotesize{2.37E-04} & \footnotesize{4.48E-04} \\[-2pt]
            & \small{Eigen Cent.} & \footnotesize{1.50E-04} & \footnotesize{3.78E-04} & \footnotesize{2.66E-05} & \footnotesize{6.56E-05} & \footnotesize{1.40E-04} \\[-2pt]
        \midrule
        \multirow{7}[-2]{*}{\begin{sideways}CoAID\end{sideways}}
            & \small{\#Nodes} & \footnotesize{48.07} & \footnotesize{94.01} & \footnotesize{7} & \footnotesize{18} & \footnotesize{44} \\[-2pt]
            & \small{\#Edges} & \footnotesize{525.98} & \footnotesize{1717.85} & \footnotesize{24} & \footnotesize{80.5} & \footnotesize{257.75} \\[-2pt]
            & \small{\#Docs.} & \footnotesize{1.86} & \footnotesize{1.58} & \footnotesize{1} & \footnotesize{1} & \footnotesize{2} \\[-2pt]
            & \small{Density} & \footnotesize{0.74} & \footnotesize{0.6} & \footnotesize{0.27} & \footnotesize{0.56} & \footnotesize{1.07} \\[-2pt]
            & \small{Prop. Docs.} & \footnotesize{13.93\%} & \footnotesize{14.36\%} & \footnotesize{2.93\%} & \footnotesize{8.33\%} & \footnotesize{20.00\%} \\[-2pt]
            & \small{Deg. Cent.} & \footnotesize{1.67E-03} & \footnotesize{2.73E-03} & \footnotesize{6.02E-04} & \footnotesize{8.02E-04} & \footnotesize{1.60E-03} \\[-2pt]
            & \small{Eigen Cent.} & \footnotesize{5.47E-04} & \footnotesize{4.44E-03} & \footnotesize{1.44E-05} & \footnotesize{4.83E-05} & \footnotesize{1.32E-04} \\[-2pt]
        \midrule
        \multirow{7}[-2]{*}{\begin{sideways}TwitterHS\end{sideways}}
            & \small{\#Nodes} & \footnotesize{1485.73} & \footnotesize{1009.45} & \footnotesize{220} & \footnotesize{2056} & \footnotesize{2571} \\[-2pt]
            & \small{\#Edges} & \footnotesize{4511.72} & \footnotesize{2967.31} & \footnotesize{1774} & \footnotesize{6168} & \footnotesize{7711} \\[-2pt]
            & \small{\#Docs.} & \footnotesize{1476.97} & \footnotesize{1017.5} & \footnotesize{22} & \footnotesize{2053} & \footnotesize{2569} \\[-2pt]
            & \small{Density} & \footnotesize{0.23} & \footnotesize{0.49} & \footnotesize{0} & \footnotesize{0} & \footnotesize{0.13} \\[-2pt]
            & \small{Prop. Docs.} & \footnotesize{81.80\%} & \footnotesize{32.20\%} & \footnotesize{81.25\%} & \footnotesize{99.85\%} & \footnotesize{99.92\%} \\[-2pt]
            & \small{Deg. Cent.} & \footnotesize{2.97E-04} & \footnotesize{1.08E-19} & \footnotesize{2.97E-04} & \footnotesize{2.97E-04} & \footnotesize{2.97E-04} \\[-2pt]
            & \small{Eigen Cent.} & \footnotesize{4.12E-03} & \footnotesize{6.36E-03} & \footnotesize{7.01E-08} & \footnotesize{3.70E-07} & \footnotesize{1.39E-02} \\[-2pt]
        \bottomrule
    \end{tabular}%
    \caption{Same as Table \ref{tab:ubs_support_stats}, but now for the query graphs.}
    \label{tab:ubs_query_stats}
\end{table*}

\begin{table*}[t]
    \centering
    \resizebox{\textwidth}{!}{
    \begin{tabular}{lcccccc}
        \toprule
        \multicolumn{1}{l}{\textbf{Parameter}} & \textbf{Full} & \textbf{Subgraphs} & \textbf{MAML-LH} & \textbf{MAML-RH} & \textbf{ProtoNet} & \textbf{ProtoMAML} \\
        \midrule
        GAT Hidden Dim & \multicolumn{6}{c}{256} \\
        GAT Heads & \multicolumn{6}{c}{3} \\
        CLF Dim & \multicolumn{6}{c}{64} \\
        \midrule
        \multicolumn{7}{c}{\textbf{Training \& Adaptation}} \\
        \midrule
        Dropout & 0.5   & 0.4   & 0.5   & 0.5   & 0.5   & 0.5 \\
        Node Dropout & 0.1   & 0.1   & 0.1   & 0.1   & 0.1   & 0.1 \\
        Attn. Dropout & 0.1   & 0.0   & 0.1   & 0.1   & 0.1   & 0.1 \\
        LR & 2.50E-03 & 2.50E-03 & 5.00E-04 & 1.00E-03 & 1.00E-03 & 1.00E-03 \\
        Weight Decay & 1.00E-02 & 5.00E-02 & 5.00E-02 & 5.00E-02 & 5.00E-02 & 5.00E-02 \\
        Batch size & N/A   & 32    & N/A   & N/A   & N/A   & N/A \\
        Updates & 100   & 300   & 2560  & 2560  & 2560  & 2560 \\
        Decay Updates & 5     & 15    & 128   & 128   & 128   & 128 \\
        Decay Factor & \multicolumn{6}{c}{0.7943} \\
        Patience & 10    & 30    & 256   & 256   & 256   & 256 \\
        \midrule
        \multicolumn{7}{c}{\textbf{Inner Loop Adaptation - Training}} \\
        \midrule
        LR - GAT & N/A   & N/A   & 1.00E-03 & 1.00E-02 & N/A   & 1.00E-02 \\
        LR - CLF Head & N/A & N/A & 1.00E-03 & 5.00E-02 & N/A & 5.00E-02 \\
        $T_{\text{inner}}$ & N/A & N/A & 1     & 5     & N/A & 10 \\
        \midrule
        \multicolumn{7}{c}{\textbf{Inner Loop Adaptation - High Adaptation Evaluation}} \\
        \midrule
        LR - GAT & N/A   & 1.00E-02   & 5.00E-03 & 5.00E-02 & N/A   & 1.00E-02 \\
        LR - CLF Head & N/A & 5.00E-01 & 5.00E-02 & 5.00E-02 & N/A & 5.00E-01 \\
        $T_{\text{inner}}$ & N/A & 25 & 25 & 25 & N/A & 25 \\
        \bottomrule
    \end{tabular}}
    \caption{Hyperparameters used for pre-training models on GossipCop, and the high-adaptation evaluation during the transfer experiments.}
    \label{tab:gossipcop_parameters}
\end{table*}
  
\section{Hyperparameters}
\label{appendix:hyperparameters}
\subsection{GossipCop Training}

The set of hyperparameters used for pre-training meta-learners on GossipCop are presented in Table \ref{tab:gossipcop_parameters}. The model size was left constant. Input dimensions were determined by RoBERTa, and set to 768. The GAT attention heads, 3, used an internal dimensionality of 256, and all heads were concatenated afterwards. After GAT processing, representations were fed through a two layer ReLU activated MLP of dimensionality 64, before being classified.

All other hyperparameters were tuned on the validation graphs generated by inductive stratified 5-fold cross validation. Dropout was applied on the internal hidden dimensions. Dropout was applied node-wise on the initial node embeddings, stochastically setting entire nodes to 0. In our case specifically, this essentially means converting documents into users. Attention dropout was applied to the attention weights produced by the GAT layers (Equation \ref{eq:gat_aggregation}). Alternatively, we also experimented with \textit{node masking}, but we did not see a big difference in performance. \cite{node_masking} AdamW \cite{kingmaAdamMethodStochastic2017,loshchilovDecoupledWeightDecay2019} was the outer-loop optimizer, with only the learning and weight decay terms tuned. The different algorithms required substantially different numbers of gradient update before convergence. Early stopping was used, with patience equal to 10\% of the maximum allowed number of steps. Most checkpoints converged well before that point. The learning rate was decayed in a step-wise manner, every 5\% of the maximum number of steps, with a minimum learning rate at 0.01 of the initial value.

The inner-loop learning rate saw more variation between the different learning algorithms. \textsc{maml-lh} performs best under minimal adaptation, yielding a single step inner loop with a low learning rate. Resetting the head in each episode, instead, forces adaptation, reflected in a larger, more aggressive inner loop. \textsc{protomaml}, finally, reaches minimum validation loss only with large amounts of inner-loop adaptation. 

During generalisation to unseen graphs, a more aggressive adaptation strategy was applied to most models. This was not tuned on the test set. Instead, the highest values possible were, such that no infinities appeared in the output logits.

\subsection{Additional Training Details}

All experiments were conducted on a Linux-based SLURM-based academic cluster. Nodes consisted of am Intel Xeon Platinum 8360Y CPU with 18 cores in user at 2.4 GHz, a single NVIDIA A100 GPU accelerator (yielding 40 GiB of HMB2 memory) and 128 GiB of DDR4 memory. The code is written exclusively using Python 3.10.6, PyTorch 1.13.0, built with CUDA 11.7. Graph modelling utilized PyTorch Geometric 2.3.0. All experiments were conducted under random seed 942. For local development we use Ubuntu 20.04.6 LTS (GNU/Linux 5.15.90.1-microsoft-standard-WSL2 x86\_64).

\section{Comparing \textsc{maml-lh} \& \textsc{maml-rh}}
\label{appendix:learned_and_reset_head}

\begin{figure*}[t]
      \includegraphics[width=\textwidth]{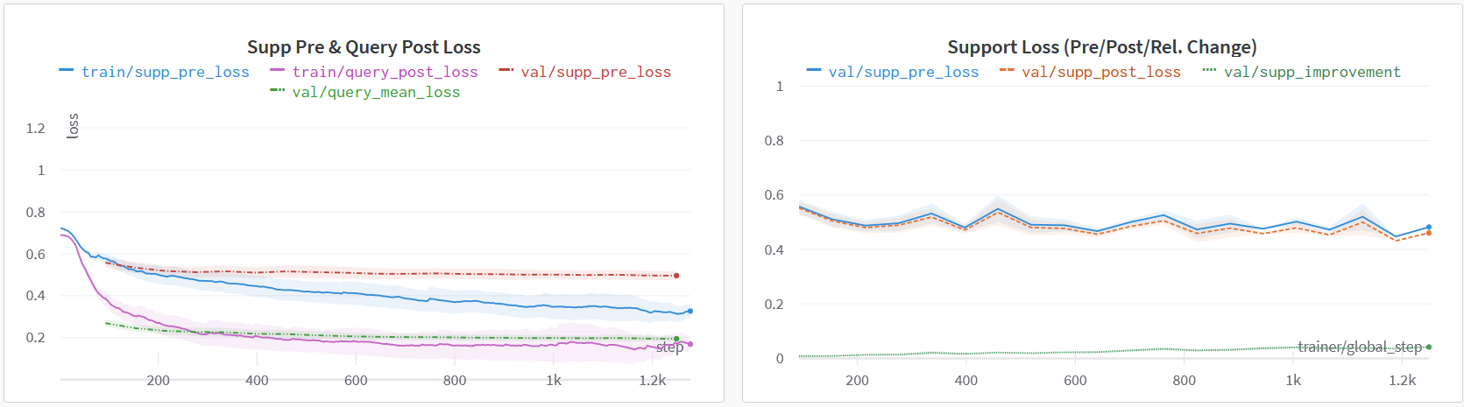}
      \includegraphics[width=\textwidth]{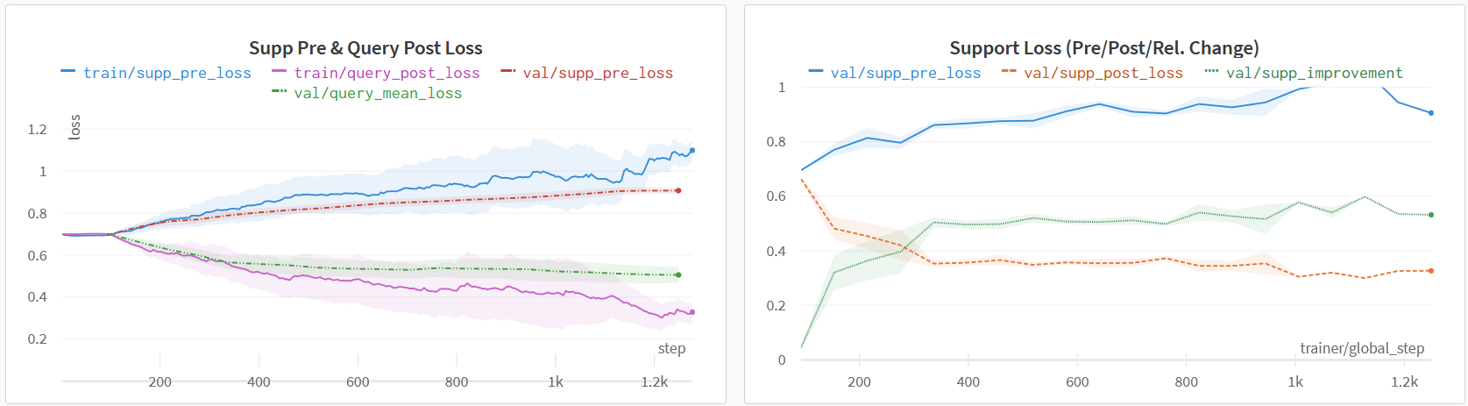}
      \caption{GossipCop training losses for (t) \textsc{maml-lh} and (b) \textsc{maml-rh}. The left column gives loss of the models on the support and query sets. Support loss is computed prior to the first adaptation step (blue and dashed orange lines), query loss after the last adaptation step (pink and dashed green lines). The left column provides the support loss prior to the first adaptation step (blue line) and after the last adaptation step (orange line). Finally, the green line gives the relative improvement of the support set loss.}
      \label{fig:learned_vs_random_head_velocity}
\end{figure*}

As originally described in \citet{finn_ModelAgnosticMetaLearningFast2017}, MAML resets its classification head each episode in order to adapt to a new task with new labels. However, in our case, pre-training MAML uses only a single task, with a fixed label definition. Therefore, the classification head can be learned in the outer loop along with the other meta-initialized models. This setting we dubbed \textsc{maml-lh} (learned head), and has the benefit of requiring less inner-loop adaptation; at least the classifier head is already task-specific. The standard MAML setup we dubbed \textsc{maml-rh} (random head).

On GossipCop, at least, this had a dramatic effect on the degree of adaptation, and as a result, on the final performance scores. This can be seen in Figure \ref{fig:learned_vs_random_head_velocity}, with the top row of figures giving various losses for \textsc{maml-lh}, and the bottom row for \textsc{maml-rh}. The left column of figures present the loss on the support set (prior to adaptation), along with the loss on the query set (post adaptation), on both the train and validation splits. \textsc{maml-lh} acts like a standard machine learnig model. After random performance initially, train loss decreases steadily on both graphs, whereas validation loss stagnates earlier on. The query loss is lower, but as we're using foMAML with a disjoint support--query split this is to be expected (the model is never directly optimised on the support graphs). \textsc{maml-rh}, on the other hand, shows rapid divergence in the support loss, while the support loss decreases as usual.

These loss patterns indicate a distinction between the two operating modes of MAML trained models. \textsc{maml-lh}, seeing a stable learning objective, learns to initialise using representations optimal to all tasks. In the meta-learning literature this corresponds to `feature reuse', and makes \textsc{maml-lh} similar to the ANIL \cite{raghuRapidLearningFeature2020} MAML variant. \textsc{maml-rh}, on the other hand, has to leverage the support set to rapdily adapt in order to achieve non-random performance on the query set. Its initial weights are not usable for representation learning, but rather for optimizing itself into a representation learner. This phenomenon is called `rapid learning'. This make \textsc{maml-rh} more reminiscent of `true' MAML, or the BOIL variant \cite{ohBOILRepresentationChange2021}.

This is made more clear in the right column of Figure \ref{fig:learned_vs_random_head_velocity}. Here the support loss before and after adaptation is shown, with a green line also indicating the relative decrease. For \textsc{maml-lh}, there is barely any difference between the two, but the overall line is already relatively low; it simply does not need to adapt to achieve generalisble representations. \textsc{maml-rh}, again, is a polar opposite, with diverging initial support loss, but a much lower final loss. The relative improvement is indicative of a model that `learns-to-learn'.

In order to test which meta-learning property is more important for the task at hand, both were trained and applied. While \textsc{maml-lh} clearly showed itself superior on GossipCop, we initially thought that added bias toward rapid adaptation might aid generalisation with \textsc{maml-rh}.

\section{Homophily}
\label{appendix:homophily}

\begin{figure*}[p]
      \includegraphics[width=\textwidth, height=0.9\textheight]{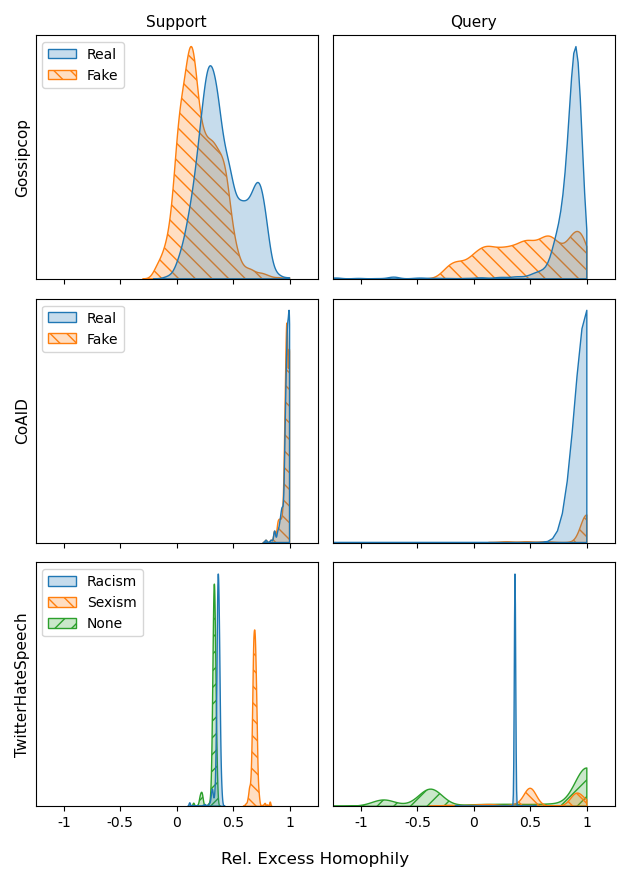}
      \caption{Kernel density estimates for the distribution of relative excess homophily (Equation \ref{eq:rel_excess_homophily}) for sampled subgraphs. The left column present user-centred sampled graphs, the right column gives the $r$-radius neighbourhoods about document nodes from the query graph. The different rows give different datasets. On the x-axis, 0 corresponds to a random graph, 1 to a perfectly homophilic graph. Values below 0 indicate heterophily.}
      \label{fig:homophily_distribution}
\end{figure*}

Graph homophily is the property of nodes to prefer attaching to similar nodes. In social network users, where interaction usually denotes some form of kinship, this follows naturally from people's social relationships. In malicious content detection, this is likely a relevant feature as well. Our perception of real and fake news is influenced by our social network neighbourhood, and propagation usually occurs in homophilic settings \cite{sunSharingNewsOnline2022}.

The effect of homophily on GNN performance remains an open question. In a homophilic setting, node representations will be built from nodes of the same class, whereas in heterophilic settings, node representations contains representations of nodes of different classes. A third setting, less explored, is the case of randomness: nodes are just as likely to attach to nodes of a different class as its own. \citet{zhuHomophilyGraphNeural2020} define a homophily metric that measures the global propensity of links between similar nodes. They show that GNNs can fail in heterophilic settings, with an MLP being more effective, despite the graph setting. For a graph $\mathcal{G}=(\mathcal{V}, \mathcal{E})$, they define homophily as,
\begin{equation}
      h^{\text{(edge)}}=\dfrac{|\{(u,v)| (u,v)\in\mathcal{E}\vee y_{u}=y_{v}\}|}{|\mathcal{E}|},
\end{equation}
i.e., the ratio of edges from nodes to similarly labelled nodes, to all edges.

Later papers dispute the claim that GNNs cannot perform under heterophily. \citet{maHomophilyNecessityGraph2021} find that GNNs require certain conditions to be met for class separation (used as a proxy for classification performance). Specifically, they indicate that as long as nodes of the same label share similar neighbourhood patterns, node representations will become more similar, despite dissimilar neighbouring nodes.

\citet{limLargeScaleLearning2021} take issue with the definition of the homophily metric. Graphs with many node labels will naturally be less homophilic. They propose a metric that measures homophily while correcting for a randomly connected null model where nodes. Extending to neighbourhoods of radius $r$,
\begin{align*}
      h^{\text{(class insen.)}}_{r}=&\frac{1}{|C|-1}\sum_{c=1}^{|C|}\lfloor h_{c}^{\text{(neigh.)}}-p_{c}\rfloor_{+}, \numberthis\label{eq:excess_edge_homophily} \\
      h_{r,c}^{\text{(neigh.)}}&=\frac{\sum_{v\in \mathcal{V}_{c}}|\{u|\mathcal{N}_{r}(v)\wedge y_{u}=y_{v}\}|}{\sum_{v\in \mathcal{V}_{c}}|\{u|\mathcal{N}_{r}(v)\}|}, \\
      &p_{c}=\frac{|\mathcal{V}_{c}|}{|\mathcal{V}|}.
\end{align*}
It may be interpreted as measuring the expected \textit{excess} homophily present in neighbourhoods about nodes of class $c$.

The proposed metrics for measuring homophily summarize whole graphs, make no distinction between user or document nodes, and do differentiate between a randomly connected graph or a heterophilic graph. In the proposed method of this paper, a homophily metric must be comparable across many subgraphs. As such, inspired by the measure of assortativity, introduced by \citet{newmanMixingPatternsNetworks2003}, we slightly modify the homophily definition as,
\begin{align*}
      \hat{h}^{\text{(subgraph)}}_{r,c}&=\frac{1}{|\mathcal{V}_{c}|}\sum_{v\in\mathcal{V}_{c}}\frac{h_{r,c}^{\text{(neigh.)}}(v)-p_c}{1-p_c}, \numberthis \label{eq:rel_excess_homophily}\\
      h_{r,c}^{\text{(neigh.)}}(v)&=\frac{|\{u|\mathcal{N}^{\text{(docs)}}_{r}(v)\wedge y_{u}=y_{v}\}|}{|\{u|\mathcal{N}^{\text{(docs)}}_{r}(v)\}|}.
\end{align*}
For a subgraph, it defines the homophily of class $c$ as the expected ratio of homophilic nodes in the $r$ radius neighbourhood of nodes $v\in\mathcal{V}_{c}$, in excess of a random graph. The use of a neighbourhood to compute $h_{r,c}^{\text{(neigh.)}}$ is deliberate. It can now measure the effect of other document nodes on the representation of the centre node. The division by $1-p_{c}$ normalizes the excess: a score of 1 is achieved only if fully homophilic, 0 if random, and $-\dfrac{p_{c}}{1-p_{c}}$ if perfectly heterophilic. This allows for interpreting homophily on a scale.  

More important, Equation \ref{eq:rel_excess_homophily} is applicable for support graphs (which have multiple nodes of each label), and query graphs (which have a single node from a single label). For the support graph, nodes' homophily scores are averaged, per class, to produce a single summary statistic. For the query graphs, the non-labelled nodes scores are simply omitted.

Rather than presenting a single summary metric, the distribution of homophilic nodes can be observed for sampled subgraphs. This is presented in Figure \ref{fig:homophily_distribution}. The left column presents the distribution of homophily scores for support graphs, right the query graphs. The rows present the different datasets.

For GossipCop, the support graphs show a relatively broad distribution of homophily scores, with modes between the 0-0.5 range. The query graph, however, shows a significant difference between the two classes, with the real documents being highly homophilic, and fake less so. In other words, there are fake documents whose neighbourhood consists primarily of real documents. CoAID shows more consistent behaviour, with both node classes being extremely homophilic.

TwitterHateSpeech is somewhat of an outlier, with substantial differences between the classes. Again, the collection procedure used by \citeauthor{waseemHatefulSymbolsHateful2016} led to a small number of extremely active, racist users. Furthermore, user-document links represent authorship, not tweet/re-tweet interactions. As a result, document representations are built up entirely out of the \textit{other} posts by the same user. As a result, the racist class has a narrow distribution, that indicates slight homophily. The innocuous class, `None', gives a bimodal distribution in the query graphs. Many innocuous documents seem to be produced by regular Twitter users, whereas a large portion come from racist and sexist users (a large bump in the heterophilic range). Only the sexist class seems to be consistently homophilic. As a result, this dataset is relatively noisy, with representations being influenced by dissimilar documents.

\section{Extended Generalisation Results}
\label{appendix:extended_transfer}

\subsection{CoAID}
\label{appendix:coaid_transfer}

The results presented in Figure \ref{fig:transfer_results} are presented in tabular form in Tables \ref{tab:coaid_low_adapt} and \ref{tab:coaid_high_adapt}. Since CoAID matches, approximately, the pre-training task used in GossipCop, models were initially adaptated using the same classification head (in case of \textsc{subgraphs} and \textsc{maml-lh}) and inner-loop learning parameters. This corresponds to the low-adaptation setting, presented in Table \ref{tab:coaid_low_adapt}. This setting estimates direct domain transfer, much like testing the Subgraphs model at $k=0$. Performance proved disappointing for most models, with the most aggressive adapter (\textsc{protomaml}), clearly exceeding all other tested models. 

Therefore, we conducted a second round of experiments with similarly aggressive inner-loop learning, presented in Table \ref{tab:coaid_high_adapt}. The only models exempted, were \textsc{subgraphs} at $k=0$ and the \textsc{protonet}, as neither adapts. All models benefited from the more aggressive inner loop, indicating that the generalisation is not trivial. All-in-all, the highest achieved MCC was $0.1709$, for \textsc{protomaml} at $k=8$, corresponding to an F1-Fake of $0.1841$. While low relative to other F1 scores reported, this should be compared to a class prevalence of ~5\%. 

\begin{table*}[htbp]
    \centering
      \begin{tabular}{clccc}
      \toprule
      \multirow{2}[4]{*}{\textbf{k}} & \multicolumn{1}{c}{\multirow{2}[4]{*}{\textbf{Method}}} & \multicolumn{2}{c}{\textbf{F1}} & \multirow{2}[4]{*}{\textbf{MCC}} \\
        \cmidrule(l{4pt}r{4pt}){3-4}
        &       & \textbf{Real} & \textbf{Fake} & \\
      \midrule
        0 & \textsc{subgraphs} & \makecell{\small{0.2445}\\[-5pt] \footnotesize{(0.2445, 0.2446)}} & \makecell{\small{0.1121}\\[-5pt] \footnotesize{(0.112, 0.1122)}} & \makecell{\small{0.0398}\\[-5pt] \footnotesize{(0.0394, 0.0402)}} \\
      \midrule
      \multirow{5}[11]{*}{4}
        & \textsc{subgraphs} & \makecell{\small{0.7164}\\[-5pt] \footnotesize{(0.71, 0.7227)}} & \makecell{\small{0.1306}\\[-5pt] \footnotesize{(0.129, 0.1321)}} & \makecell{\small{0.0642}\\[-5pt] \footnotesize{(0.0628, 0.0656)}} \\
        & \textsc{maml-lh} & \makecell{\small{0.4225}\\[-5pt] \footnotesize{(0.4224, 0.4227)}} & \makecell{\small{0.1162}\\[-5pt] \footnotesize{(0.1161, 0.1164)}} & \makecell{\small{0.0593}\\[-5pt] \footnotesize{(0.0592, 0.0593)}} \\
        & \textsc{maml-rh} & \makecell{\small{0.8241}\\[-5pt] \footnotesize{(0.8198, 0.8285)}} & \makecell{\small{0.1561}\\[-5pt] \footnotesize{(0.1548, 0.1575)}} & \makecell{\small{0.1164}\\[-5pt] \footnotesize{(0.1149, 0.1179)}} \\
        & \textsc{protonet} & \makecell{\small{0.7007}\\[-5pt] \footnotesize{(0.6928, 0.7085)}} & \makecell{\small{0.1404}\\[-5pt] \footnotesize{(0.1388, 0.1421)}} & \makecell{\small{0.0731}\\[-5pt] \footnotesize{(0.0692, 0.0771)}} \\
        & \textsc{protomaml} & \makecell{\small{0.7867}\\[-5pt] \footnotesize{(0.7791, 0.7942)}} & \makecell{\small{0.1767}\\[-5pt] \footnotesize{(0.1746, 0.1788)}} & \makecell{\small{0.1321}\\[-5pt] \footnotesize{(0.1279, 0.1363)}} \\
      \midrule
      \multirow{5}[11]{*}{8}
        & \textsc{subgraphs} & \makecell{\small{0.2219}\\[-5pt] \footnotesize{(0.2218, 0.222)}} & \makecell{\small{0.1044}\\[-5pt] \footnotesize{(0.1043, 0.1046)}} & \makecell{\small{0.0377}\\[-5pt] \footnotesize{(0.0372, 0.0383)}} \\
        & \textsc{maml-lh} & \makecell{\small{0.5693}\\[-5pt] \footnotesize{(0.5691, 0.5695)}} & \makecell{\small{0.1096}\\[-5pt] \footnotesize{(0.1094, 0.1098)}} & \makecell{\small{0.0571}\\[-5pt] \footnotesize{(0.057, 0.0572)}} \\
        & \textsc{maml-rh} & \makecell{\small{0.8243}\\[-5pt] \footnotesize{(0.8208, 0.8278)}} & \makecell{\small{0.1567}\\[-5pt] \footnotesize{(0.1556, 0.1577)}} & \makecell{\small{0.1234}\\[-5pt] \footnotesize{(0.1221, 0.1247)}} \\
        & \textsc{protonet} & \makecell{\small{0.7540}\\[-5pt] \footnotesize{(0.7493, 0.7588)}} & \makecell{\small{0.1471}\\[-5pt] \footnotesize{(0.1458, 0.1484)}} & \makecell{\small{0.1179}\\[-5pt] \footnotesize{(0.1154, 0.1205)}} \\
        & \textsc{protomaml} & \makecell{\small{0.8300}\\[-5pt] \footnotesize{(0.826, 0.8341)}} & \makecell{\small{0.1799}\\[-5pt] \footnotesize{(0.1784, 0.1814)}} & \makecell{\small{0.1616}\\[-5pt] \footnotesize{(0.1594, 0.1638)}} \\
      \midrule
      \multirow{5}[11]{*}{12}
        & \textsc{subgraphs} & \makecell{\small{0.2387}\\[-5pt] \footnotesize{(0.2386, 0.2389)}} & \makecell{\small{0.0974}\\[-5pt] \footnotesize{(0.0972, 0.0975)}} & \makecell{\small{0.0370}\\[-5pt] \footnotesize{(0.0363, 0.0377)}} \\
        & \textsc{maml-lh} & \makecell{\small{0.5740}\\[-5pt] \footnotesize{(0.5738, 0.5741)}} & \makecell{\small{0.1028}\\[-5pt] \footnotesize{(0.1025, 0.103)}} & \makecell{\small{0.0551}\\[-5pt] \footnotesize{(0.055, 0.0552)}} \\
        & \textsc{maml-rh} & \makecell{\small{0.8355}\\[-5pt] \footnotesize{(0.8328, 0.8383)}} & \makecell{\small{0.1466}\\[-5pt] \footnotesize{(0.1457, 0.1475)}} & \makecell{\small{0.1154}\\[-5pt] \footnotesize{(0.1142, 0.1167)}} \\
        & \textsc{protonet} & \makecell{\small{0.7560}\\[-5pt] \footnotesize{(0.7517, 0.7604)}} & \makecell{\small{0.1395}\\[-5pt] \footnotesize{(0.1384, 0.1405)}} & \makecell{\small{0.1175}\\[-5pt] \footnotesize{(0.1154, 0.1195)}} \\
        & \textsc{protomaml} & \makecell{\small{0.8331}\\[-5pt] \footnotesize{(0.8295, 0.8367)}} & \makecell{\small{0.1675}\\[-5pt] \footnotesize{(0.1662, 0.1687)}} & \makecell{\small{0.1583}\\[-5pt] \footnotesize{(0.1566, 0.1601)}} \\
      \midrule
      \multirow{5}[11]{*}{16}
        & \textsc{subgraphs} & \makecell{\small{0.2296}\\[-5pt] \footnotesize{(0.2294, 0.2297)}} & \makecell{\small{0.0896}\\[-5pt] \footnotesize{(0.0894, 0.0898)}} & \makecell{\small{0.0379}\\[-5pt] \footnotesize{(0.0371, 0.0387)}} \\
        & \textsc{maml-lh} & \makecell{\small{0.5871}\\[-5pt] \footnotesize{(0.5869, 0.5873)}} & \makecell{\small{0.0956}\\[-5pt] \footnotesize{(0.0954, 0.0959)}} & \makecell{\small{0.0528}\\[-5pt] \footnotesize{(0.0527, 0.0529)}} \\
        & \textsc{maml-rh} & \makecell{\small{0.8315}\\[-5pt] \footnotesize{(0.8291, 0.834)}} & \makecell{\small{0.1352}\\[-5pt] \footnotesize{(0.1343, 0.1361)}} & \makecell{\small{0.1106}\\[-5pt] \footnotesize{(0.1093, 0.112)}} \\
        & \textsc{protonet} & \makecell{\small{0.7429}\\[-5pt] \footnotesize{(0.7383, 0.7475)}} & \makecell{\small{0.1268}\\[-5pt] \footnotesize{(0.1259, 0.1278)}} & \makecell{\small{0.1097}\\[-5pt] \footnotesize{(0.1079, 0.1115)}} \\
        & \textsc{protomaml} & \makecell{\small{0.8323}\\[-5pt] \footnotesize{(0.8291, 0.8355)}} & \makecell{\small{0.1530}\\[-5pt] \footnotesize{(0.1518, 0.1541)}} & \makecell{\small{0.1508}\\[-5pt] \footnotesize{(0.1493, 0.1523)}} \\
    \bottomrule
    \end{tabular}%
    \caption{CoAID transfer results under low adaptation hyperparameters.}
    \label{tab:coaid_low_adapt}
\end{table*}

\begin{table*}[htbp]
    \centering
      \begin{tabular}{clccc}
      \toprule
      \multirow{2}[4]{*}{\textbf{k}} & \multicolumn{1}{c}{\multirow{2}[4]{*}{\textbf{Method}}} & \multicolumn{2}{c}{\textbf{F1}} & \multirow{2}[4]{*}{\textbf{MCC}} \\
        \cmidrule(l{4pt}r{4pt}){3-4}
        &       & \textbf{Real} & \textbf{Fake} & \\
    \midrule
        0     & \textsc{subgraphs} & \makecell{\small{0.2445}\\[-5pt] \footnotesize{(0.2445, 0.2446)}} & \makecell{\small{0.1121}\\[-5pt] \footnotesize{(0.112, 0.1122)}} & \makecell{\small{0.0398}\\[-5pt] \footnotesize{(0.0394, 0.0402)}} \\
    \midrule
    \multirow{5}[11]{*}{4}
        & \textsc{subgraphs} & \makecell{\small{0.727}\\[-5pt] \footnotesize{(0.7199, 0.7341)}} & \makecell{\small{0.131}\\[-5pt] \footnotesize{(0.1289, 0.1331)}} & \makecell{\small{0.0727}\\[-5pt] \footnotesize{(0.0708, 0.0746)}} \\
        & \textsc{maml-lh} & \makecell{\small{0.7494}\\[-5pt] \footnotesize{(0.7435, 0.7553)}} & \makecell{\small{0.1538}\\[-5pt] \footnotesize{(0.1524, 0.1552)}} & \makecell{\small{0.1172}\\[-5pt] \footnotesize{(0.1154, 0.119)}} \\
        & \textsc{maml-rh} & \makecell{\small{0.7797}\\[-5pt] \footnotesize{(0.7744, 0.7851)}} & \makecell{\small{0.1562}\\[-5pt] \footnotesize{(0.1547, 0.1577)}} & \makecell{\small{0.1195}\\[-5pt] \footnotesize{(0.1177, 0.1214)}} \\
        & \textsc{protonet} & \makecell{\small{0.7007}\\[-5pt] \footnotesize{(0.6928, 0.7085)}} & \makecell{\small{0.1404}\\[-5pt] \footnotesize{(0.1388, 0.1421)}} & \makecell{\small{0.0731}\\[-5pt] \footnotesize{(0.0692, 0.0771)}} \\
        & \textsc{protomaml} & \makecell{\small{0.7734}\\[-5pt] \footnotesize{(0.7655, 0.7812)}} & \makecell{\small{0.1762}\\[-5pt] \footnotesize{(0.174, 0.1784)}} & \makecell{\small{0.1383}\\[-5pt] \footnotesize{(0.1343, 0.1422)}} \\
    \midrule
    \multirow{5}[11]{*}{8}
        & \textsc{subgraphs} & \makecell{\small{0.7651}\\[-5pt] \footnotesize{(0.7597, 0.7705)}} & \makecell{\small{0.129}\\[-5pt] \footnotesize{(0.1268, 0.1312)}} & \makecell{\small{0.0827}\\[-5pt] \footnotesize{(0.0807, 0.0846)}} \\
        & \textsc{maml-lh} & \makecell{\small{0.7679}\\[-5pt] \footnotesize{(0.7633, 0.7724)}} & \makecell{\small{0.1525}\\[-5pt] \footnotesize{(0.1512, 0.1537)}} & \makecell{\small{0.1232}\\[-5pt] \footnotesize{(0.1215, 0.1248)}} \\
        & \textsc{maml-rh} & \makecell{\small{0.8021}\\[-5pt] \footnotesize{(0.7983, 0.8059)}} & \makecell{\small{0.1581}\\[-5pt] \footnotesize{(0.157, 0.1593)}} & \makecell{\small{0.1312}\\[-5pt] \footnotesize{(0.1297, 0.1326)}} \\
        & \textsc{protonet} & \makecell{\small{0.754}\\[-5pt] \footnotesize{(0.7493, 0.7588)}} & \makecell{\small{0.1471}\\[-5pt] \footnotesize{(0.1458, 0.1484)}} & \makecell{\small{0.1179}\\[-5pt] \footnotesize{(0.1154, 0.1205)}} \\
        & \textsc{protomaml} & \makecell{\small{0.8245}\\[-5pt] \footnotesize{(0.8202, 0.8288)}} & \makecell{\small{0.1841}\\[-5pt] \footnotesize{(0.1824, 0.1858)}} & \makecell{\small{0.1709}\\[-5pt] \footnotesize{(0.1685, 0.1733)}} \\
    \midrule
    \multirow{5}[11]{*}{12}
        & \textsc{subgraphs} & \makecell{\small{0.7783}\\[-5pt] \footnotesize{(0.774, 0.7825)}} & \makecell{\small{0.1316}\\[-5pt] \footnotesize{(0.1297, 0.1334)}} & \makecell{\small{0.0921}\\[-5pt] \footnotesize{(0.0904, 0.0939)}} \\
        & \textsc{maml-lh} & \makecell{\small{0.7893}\\[-5pt] \footnotesize{(0.7856, 0.7931)}} & \makecell{\small{0.147}\\[-5pt] \footnotesize{(0.1459, 0.1482)}} & \makecell{\small{0.1263}\\[-5pt] \footnotesize{(0.1248, 0.1279)}} \\
        & \textsc{maml-rh} & \makecell{\small{0.8181}\\[-5pt] \footnotesize{(0.8149, 0.8213)}} & \makecell{\small{0.1505}\\[-5pt] \footnotesize{(0.1495, 0.1516)}} & \makecell{\small{0.1281}\\[-5pt] \footnotesize{(0.1267, 0.1295)}} \\
        & \textsc{protonet} & \makecell{\small{0.756}\\[-5pt] \footnotesize{(0.7517, 0.7604)}} & \makecell{\small{0.1395}\\[-5pt] \footnotesize{(0.1384, 0.1405)}} & \makecell{\small{0.1175}\\[-5pt] \footnotesize{(0.1154, 0.1195)}} \\
        & \textsc{protomaml} & \makecell{\small{0.8294}\\[-5pt] \footnotesize{(0.8255, 0.8333)}} & \makecell{\small{0.1732}\\[-5pt] \footnotesize{(0.1718, 0.1746)}} & \makecell{\small{0.1689}\\[-5pt] \footnotesize{(0.1669, 0.1708)}} \\
    \midrule
    \multirow{5}[11]{*}{16}
        & \textsc{subgraphs} & \makecell{\small{0.7922}\\[-5pt] \footnotesize{(0.7887, 0.7956)}} & \makecell{\small{0.1314}\\[-5pt] \footnotesize{(0.1298, 0.133)}} & \makecell{\small{0.1028}\\[-5pt] \footnotesize{(0.1011, 0.1045)}} \\
        & \textsc{maml-lh} & \makecell{\small{0.806}\\[-5pt] \footnotesize{(0.8033, 0.8087)}} & \makecell{\small{0.1385}\\[-5pt] \footnotesize{(0.1375, 0.1396)}} & \makecell{\small{0.1277}\\[-5pt] \footnotesize{(0.1263, 0.1291)}} \\
        & \textsc{maml-rh} & \makecell{\small{0.8165}\\[-5pt] \footnotesize{(0.8139, 0.8192)}} & \makecell{\small{0.1405}\\[-5pt] \footnotesize{(0.1396, 0.1414)}} & \makecell{\small{0.1234}\\[-5pt] \footnotesize{(0.1221, 0.1247)}} \\
        & \textsc{protonet} & \makecell{\small{0.7429}\\[-5pt] \footnotesize{(0.7383, 0.7475)}} & \makecell{\small{0.1268}\\[-5pt] \footnotesize{(0.1259, 0.1278)}} & \makecell{\small{0.1097}\\[-5pt] \footnotesize{(0.1079, 0.1115)}} \\
        & \textsc{protomaml} & \makecell{\small{0.8321}\\[-5pt] \footnotesize{(0.8288, 0.8354)}} & \makecell{\small{0.1599}\\[-5pt] \footnotesize{(0.1587, 0.1612)}} & \makecell{\small{0.1646}\\[-5pt] \footnotesize{(0.1631, 0.1662)}} \\
    \bottomrule
    \end{tabular}%
    \caption{CoAID transfer results under high adaptation hyperparameters, with the exception for \textsc{subgraphs} at $k=0$ and \textsc{protonet}, neither of which adapts during evaluation.}
    \label{tab:coaid_high_adapt}
\end{table*}

\subsection{TwitterHateSpeech}
\label{appendix:twitter_transfer}

Similarly, the TwitterHateSpeech results presented in Figure \ref{fig:transfer_results} are presented in tabular form in Table \ref{tab:twitter_transfer}. Having learnt from CoAID, only the high-adaptation hyperparameters were used. Lower $k$-shot values see the `None' class dominate in terms of classification scores. Performance on the minority classes increased with larger $k$-shot values, at the cost of reduced innocuous tweet performance. Ultimately, ProtoMAML manages this trade-off best, with gradually increasing MCC scores.

\begin{table*}[p]
    \centering
    \begin{tabular}{clcccc}
    \toprule
    \multirow{2}[4]{*}{\textbf{k}} & \multicolumn{1}{c}{\multirow{2}[4]{*}{\textbf{Method}}} & \multicolumn{3}{c}{\textbf{F1}} & \multirow{2}[4]{*}{\textbf{MCC}} \\
        \cmidrule(l{4pt}r{4pt}){3-5}
        &       & \textbf{Racism} & \textbf{Sexism} & \textbf{None} & \\
        \midrule
    \multirow{5}[12]{*}{4}
        & \textsc{subgraphs} & \makecell{\small{0.1615}\\[-5pt] \footnotesize{(0.1563, 0.1666)}} & \makecell{\small{0.1950}\\[-5pt] \footnotesize{(0.1894, 0.2007)}} & \makecell{\small{0.3745}\\[-5pt] \footnotesize{(0.3634, 0.3855)}} & \makecell{\small{0.0334}\\[-5pt] \footnotesize{(0.0305, 0.0363)}} \\
        & \textsc{maml-lh} & \makecell{\small{0.1696}\\[-5pt] \footnotesize{(0.1633, 0.1759)}} & \makecell{\small{0.2287}\\[-5pt] \footnotesize{(0.223, 0.2344)}} & \makecell{\small{0.3554}\\[-5pt] \footnotesize{(0.3461, 0.3646)}} & \makecell{\small{0.0580}\\[-5pt] \footnotesize{(0.0539, 0.0621)}} \\
        & \textsc{maml-rh} & \makecell{\small{0.1741}\\[-5pt] \footnotesize{(0.1682, 0.1801)}} & \makecell{\small{0.2402}\\[-5pt] \footnotesize{(0.2352, 0.2453)}} & \makecell{\small{0.3420}\\[-5pt] \footnotesize{(0.3322, 0.3519)}} & \makecell{\small{0.0543}\\[-5pt] \footnotesize{(0.0507, 0.0579)}} \\
        & \textsc{protonet} & \makecell{\small{0.1949}\\[-5pt] \footnotesize{(0.188, 0.2019)}} & \makecell{\small{0.2355}\\[-5pt] \footnotesize{(0.2299, 0.2411)}} & \makecell{\small{0.3930}\\[-5pt] \footnotesize{(0.3851, 0.401)}} & \makecell{\small{0.0784}\\[-5pt] \footnotesize{(0.0742, 0.0826)}} \\
        & \textsc{protomaml} & \makecell{\small{0.1763}\\[-5pt] \footnotesize{(0.1695, 0.1831)}} & \makecell{\small{0.2181}\\[-5pt] \footnotesize{(0.2124, 0.2238)}} & \makecell{\small{0.3585}\\[-5pt] \footnotesize{(0.3494, 0.3676)}} & \makecell{\small{0.0607}\\[-5pt] \footnotesize{(0.0565, 0.0648)}} \\
        \midrule
    \multirow{5}[12]{*}{8}
        & \textsc{subgraphs} & \makecell{\small{0.1545}\\[-5pt] \footnotesize{(0.1493, 0.1598)}} & \makecell{\small{0.2161}\\[-5pt] \footnotesize{(0.2108, 0.2214)}} & \makecell{\small{0.3750}\\[-5pt] \footnotesize{(0.3646, 0.3854)}} & \makecell{\small{0.0333}\\[-5pt] \footnotesize{(0.0304, 0.0361)}} \\
        & \textsc{maml-lh} & \makecell{\small{0.1802}\\[-5pt] \footnotesize{(0.1736, 0.1869)}} & \makecell{\small{0.2552}\\[-5pt] \footnotesize{(0.2499, 0.2606)}} & \makecell{\small{0.3340}\\[-5pt] \footnotesize{(0.3248, 0.3432)}} & \makecell{\small{0.0817}\\[-5pt] \footnotesize{(0.0781, 0.0853)}} \\
        & \textsc{maml-rh} & \makecell{\small{0.1669}\\[-5pt] \footnotesize{(0.1608, 0.173)}} & \makecell{\small{0.2614}\\[-5pt] \footnotesize{(0.2574, 0.2653)}} & \makecell{\small{0.3480}\\[-5pt] \footnotesize{(0.338, 0.3581)}} & \makecell{\small{0.0515}\\[-5pt] \footnotesize{(0.0483, 0.0546)}} \\
        & \textsc{protonet} & \makecell{\small{0.2157}\\[-5pt] \footnotesize{(0.2092, 0.2222)}} & \makecell{\small{0.2221}\\[-5pt] \footnotesize{(0.2162, 0.2281)}} & \makecell{\small{0.4065}\\[-5pt] \footnotesize{(0.3991, 0.4138)}} & \makecell{\small{0.0904}\\[-5pt] \footnotesize{(0.0862, 0.0946)}} \\
        & \textsc{protomaml} & \makecell{\small{0.1934}\\[-5pt] \footnotesize{(0.1866, 0.2002)}} & \makecell{\small{0.2148}\\[-5pt] \footnotesize{(0.2092, 0.2205)}} & \makecell{\small{0.3530}\\[-5pt] \footnotesize{(0.3439, 0.362)}} & \makecell{\small{0.0699}\\[-5pt] \footnotesize{(0.0657, 0.074)}} \\
        \midrule
    \multirow{5}[12]{*}{12}
        & \textsc{subgraphs} & \makecell{\small{0.1895}\\[-5pt] \footnotesize{(0.1838, 0.1952)}} & \makecell{\small{0.2169}\\[-5pt] \footnotesize{(0.2113, 0.2224)}} & \makecell{\small{0.3686}\\[-5pt] \footnotesize{(0.3576, 0.3795)}} & \makecell{\small{0.0515}\\[-5pt] \footnotesize{(0.0482, 0.0549)}} \\
        & \textsc{maml-lh} & \makecell{\small{0.1898}\\[-5pt] \footnotesize{(0.183, 0.1966)}} & \makecell{\small{0.2543}\\[-5pt] \footnotesize{(0.2487, 0.26)}} & \makecell{\small{0.3068}\\[-5pt] \footnotesize{(0.2976, 0.316)}} & \makecell{\small{0.0770}\\[-5pt] \footnotesize{(0.0733, 0.0807)}} \\
        & \textsc{maml-rh} & \makecell{\small{0.2286}\\[-5pt] \footnotesize{(0.2235, 0.2336)}} & \makecell{\small{0.2525}\\[-5pt] \footnotesize{(0.2468, 0.2581)}} & \makecell{\small{0.3185}\\[-5pt] \footnotesize{(0.3084, 0.3285)}} & \makecell{\small{0.0791}\\[-5pt] \footnotesize{(0.0757, 0.0824)}} \\
        & \textsc{protonet} & \makecell{\small{0.2798}\\[-5pt] \footnotesize{(0.2743, 0.2852)}} & \makecell{\small{0.2612}\\[-5pt] \footnotesize{(0.2554, 0.267)}} & \makecell{\small{0.3803}\\[-5pt] \footnotesize{(0.3731, 0.3876)}} & \makecell{\small{0.1134}\\[-5pt] \footnotesize{(0.1092, 0.1177)}} \\
        & \textsc{protomaml} & \makecell{\small{0.2545}\\[-5pt] \footnotesize{(0.2475, 0.2616)}} & \makecell{\small{0.2554}\\[-5pt] \footnotesize{(0.2492, 0.2616)}} & \makecell{\small{0.3503}\\[-5pt] \footnotesize{(0.3412, 0.3594)}} & \makecell{\small{0.1109}\\[-5pt] \footnotesize{(0.1066, 0.1152)}} \\
        \midrule
    \multirow{5}[12]{*}{16}
        & \textsc{subgraphs} & \makecell{\small{0.1996}\\[-5pt] \footnotesize{(0.194, 0.2052)}} & \makecell{\small{0.2261}\\[-5pt] \footnotesize{(0.2206, 0.2316)}} & \makecell{\small{0.3602}\\[-5pt] \footnotesize{(0.3492, 0.3712)}} & \makecell{\small{0.0574}\\[-5pt] \footnotesize{(0.054, 0.0608)}} \\
        & \textsc{maml-lh} & \makecell{\small{0.1874}\\[-5pt] \footnotesize{(0.1805, 0.1942)}} & \makecell{\small{0.2595}\\[-5pt] \footnotesize{(0.2538, 0.2652)}} & \makecell{\small{0.3072}\\[-5pt] \footnotesize{(0.2978, 0.3165)}} & \makecell{\small{0.0778}\\[-5pt] \footnotesize{(0.0741, 0.0815)}} \\
        & \textsc{maml-rh} & \makecell{\small{0.2319}\\[-5pt] \footnotesize{(0.227, 0.2368)}} & \makecell{\small{0.2510}\\[-5pt] \footnotesize{(0.2453, 0.2567)}} & \makecell{\small{0.3159}\\[-5pt] \footnotesize{(0.3059, 0.3259)}} & \makecell{\small{0.0811}\\[-5pt] \footnotesize{(0.0778, 0.0844)}} \\
        & \textsc{protonet} & \makecell{\small{0.2883}\\[-5pt] \footnotesize{(0.2838, 0.2927)}} & \makecell{\small{0.2579}\\[-5pt] \footnotesize{(0.2523, 0.2635)}} & \makecell{\small{0.3740}\\[-5pt] \footnotesize{(0.3664, 0.3816)}} & \makecell{\small{0.1128}\\[-5pt] \footnotesize{(0.1087, 0.117)}} \\
        & \textsc{protomaml} & \makecell{\small{0.3021}\\[-5pt] \footnotesize{(0.2961, 0.308)}} & \makecell{\small{0.3077}\\[-5pt] \footnotesize{(0.2999, 0.3155)}} & \makecell{\small{0.3163}\\[-5pt] \footnotesize{(0.3051, 0.3276)}} & \makecell{\small{0.1354}\\[-5pt] \footnotesize{(0.1303, 0.1404)}} \\
    \bottomrule
    \end{tabular}
    \caption{TwitterHateSpeech transfer results under high adaptation hyperparameters.}
    \label{tab:twitter_transfer}
\end{table*}

\subsection{Ablating GossipCop Pre-Training}
\label{appendix:random_transfer}

Tables \ref{tab:coaid_random_transfer} and \ref{tab:twitter_random_transfer} show additional results pertaining to the ablation experiment described in Section \ref{sec:ablation}. The largest addition, is the inclusion of the other type of GBML algorithm, MAML. The comparison model used is \textsc{maml-rh}. 

On CoAID, \textsc{maml-reset} yields `always positive' models, giving constant MCC scores of 0. ProtoMAML, however, proves reasonably robust, with \textsc{protomaml-reset} performance that exceeds trained \textsc{maml-rh}. The same effect also holds on TwitterHateSpeech, with \textsc{protomaml} only overcoming its reset counterpart in the larger $k$-shot settings.

\begin{table*}[htbp]
    \centering
    \begin{tabular}{clccc}
    \toprule
    \multirow{2}[4]{*}{\textbf{k}} & \multicolumn{1}{c}{\multirow{2}[4]{*}{\textbf{Method}}} & \multicolumn{2}{c}{\textbf{F1}} & \multirow{2}[4]{*}{\textbf{MCC}} \\
    \cmidrule(l{4pt}r{4pt}){3-4}          &       & \textbf{Real} & \textbf{Fake} &  \\
    \midrule
    \multirow{2}[2]{*}{4}\
      & \textsc{maml-reset}  & \makecell{\small{0.971}\\[-5pt] \footnotesize{(0.971, 0.9711)}} & \makecell{\small{0}\\[-5pt] \footnotesize{(0, 0)}} & \makecell{\small{0}\\[-5pt] \footnotesize{(0, 0)}} \\
      & \textsc{protomaml-reset} & \makecell{\small{0.7869}\\[-5pt] \footnotesize{(0.7674, 0.8065)}} & \makecell{\small{0.1716}\\[-5pt] \footnotesize{(0.1673, 0.1759)}} & \makecell{\small{0.1191}\\[-5pt] \footnotesize{(0.1114, 0.1268)}} \\
    \midrule
    \multirow{2}[2]{*}{8}
      & \textsc{maml-reset}  & \makecell{\small{0.9732}\\[-5pt] \footnotesize{(0.9731, 0.9732)}} & \makecell{\small{0}\\[-5pt] \footnotesize{(0, 0)}} & \makecell{\small{0}\\[-5pt] \footnotesize{(0, 0)}} \\
      & \textsc{protomaml-reset} & \makecell{\small{0.8483}\\[-5pt] \footnotesize{(0.8394, 0.8573)}} & \makecell{\small{0.1769}\\[-5pt] \footnotesize{(0.1738, 0.18)}} & \makecell{\small{0.1398}\\[-5pt] \footnotesize{(0.1361, 0.1434)}} \\
    \midrule
    \multirow{2}[2]{*}{12}
      & \textsc{maml-reset}  & \makecell{\small{0.9752}\\[-5pt] \footnotesize{(0.9751, 0.9753)}} & \makecell{\small{0}\\[-5pt] \footnotesize{(0, 0)}} & \makecell{\small{0}\\[-5pt] \footnotesize{(0, 0)}} \\
      & \textsc{protomaml-reset} & \makecell{\small{0.8470}\\[-5pt] \footnotesize{(0.8368, 0.8572)}} & \makecell{\small{0.1652}\\[-5pt] \footnotesize{(0.1621, 0.1683)}} & \makecell{\small{0.1304}\\[-5pt] \footnotesize{(0.1262, 0.1347)}} \\
    \midrule
    \multirow{2}[2]{*}{16}
      & \textsc{maml-reset}  & \makecell{\small{0.9774}\\[-5pt] \footnotesize{(0.9773, 0.9775)}} & \makecell{\small{0}\\[-5pt] \footnotesize{(0, 0)}} & \makecell{\small{0}\\[-5pt] \footnotesize{(0, 0)}} \\
      & \textsc{protomaml-reset} & \makecell{\small{0.8542}\\[-5pt] \footnotesize{(0.8479, 0.8605)}} & \makecell{\small{0.1504}\\[-5pt] \footnotesize{(0.1475, 0.1533)}} & \makecell{\small{0.1212}\\[-5pt] \footnotesize{(0.1179, 0.1245)}} \\
    \bottomrule
    \end{tabular}%
    \caption{Models `transferred' to CoAID after reset.}
    \label{tab:coaid_random_transfer}
\end{table*}

\begin{table*}[htbp]
  \centering
    \begin{tabular}{clcccc}
    \toprule
    \multirow{2}[4]{*}{\textbf{k}} & \multicolumn{1}{c}{\multirow{2}[4]{*}{\textbf{Method}}} & \multicolumn{3}{c}{\textbf{F1}} & \multirow{2}[4]{*}{\textbf{MCC}} \\
\cmidrule(l{4pt}r{4pt}){3-5}          &       & \textbf{Racism} & \textbf{Sexism} & \textbf{None} &  \\
    \midrule
    \multirow{2}[2]{*}{4} 
      & \textsc{maml-reset}  & \makecell{\small{0.1699}\\[-5pt] \footnotesize{(0.163, 0.1768)}} & \makecell{\small{0.1918}\\[-5pt] \footnotesize{(0.1853, 0.1983)}} & \makecell{\small{0.3433}\\[-5pt] \footnotesize{(0.3348, 0.3517)}} & \makecell{\small{0.0726}\\[-5pt] \footnotesize{(0.0687, 0.0765)}} \\
      & \textsc{protomaml-reset} & \makecell{\small{0.1799}\\[-5pt] \footnotesize{(0.173, 0.1868)}} & \makecell{\small{0.1906}\\[-5pt] \footnotesize{(0.1845, 0.1966)}} & \makecell{\small{0.3225}\\[-5pt] \footnotesize{(0.3136, 0.3314)}} & \makecell{\small{0.0767}\\[-5pt] \footnotesize{(0.0729, 0.0806)}} \\
    \midrule
    \multirow{2}[2]{*}{8}
      & \textsc{maml-reset}  & \makecell{\small{0.2239}\\[-5pt] \footnotesize{(0.2178, 0.2301)}} & \makecell{\small{0.1691}\\[-5pt] \footnotesize{(0.164, 0.1742)}} & \makecell{\small{0.3277}\\[-5pt] \footnotesize{(0.3193, 0.3361)}} & \makecell{\small{0.0811}\\[-5pt] \footnotesize{(0.0777, 0.0846)}} \\
      & \textsc{protomaml-reset} & \makecell{\small{0.1999}\\[-5pt] \footnotesize{(0.1934, 0.2065)}} & \makecell{\small{0.1617}\\[-5pt] \footnotesize{(0.1556, 0.1677)}} & \makecell{\small{0.3416}\\[-5pt] \footnotesize{(0.3336, 0.3495)}} & \makecell{\small{0.0868}\\[-5pt] \footnotesize{(0.0832, 0.0904)}} \\
    \midrule
    \multirow{2}[2]{*}{12}
      & \textsc{maml-reset}  & \makecell{\small{0.2315}\\[-5pt] \footnotesize{(0.2264, 0.2366)}} & \makecell{\small{0.1348}\\[-5pt] \footnotesize{(0.1292, 0.1404)}} & \makecell{\small{0.3125}\\[-5pt] \footnotesize{(0.3056, 0.3194)}} & \makecell{\small{0.0969}\\[-5pt] \footnotesize{(0.0938, 0.1001)}} \\
      & \textsc{protomaml-reset} & \makecell{\small{0.2471}\\[-5pt] \footnotesize{(0.2421, 0.2521)}} & \makecell{\small{0.1399}\\[-5pt] \footnotesize{(0.1348, 0.1449)}} & \makecell{\small{0.2882}\\[-5pt] \footnotesize{(0.28, 0.2965)}} & \makecell{\small{0.1025}\\[-5pt] \footnotesize{(0.0997, 0.1053)}} \\
    \midrule
    \multirow{2}[2]{*}{16}
      & \textsc{maml-reset}  & \makecell{\small{0.2365}\\[-5pt] \footnotesize{(0.2319, 0.241)}} & \makecell{\small{0.1294}\\[-5pt] \footnotesize{(0.1239, 0.1348)}} & \makecell{\small{0.3065}\\[-5pt] \footnotesize{(0.3003, 0.3127)}} & \makecell{\small{0.098}\\[-5pt] \footnotesize{(0.095, 0.1009)}} \\
      & \textsc{protomaml-reset} & \makecell{\small{0.2524}\\[-5pt] \footnotesize{(0.248, 0.2569)}} & \makecell{\small{0.1233}\\[-5pt] \footnotesize{(0.1184, 0.1281)}} & \makecell{\small{0.287}\\[-5pt] \footnotesize{(0.2789, 0.2952)}} & \makecell{\small{0.1052}\\[-5pt] \footnotesize{(0.1025, 0.1078)}} \\
    \bottomrule
    \end{tabular}%
    \caption{Models `transferred' to TwitterHateSpeech after reset.}
    \label{tab:twitter_random_transfer}%
\end{table*}%

\subsection{Extreme \textit{k}-shot}
\begin{figure*}[t]
      \includegraphics[width=\textwidth]{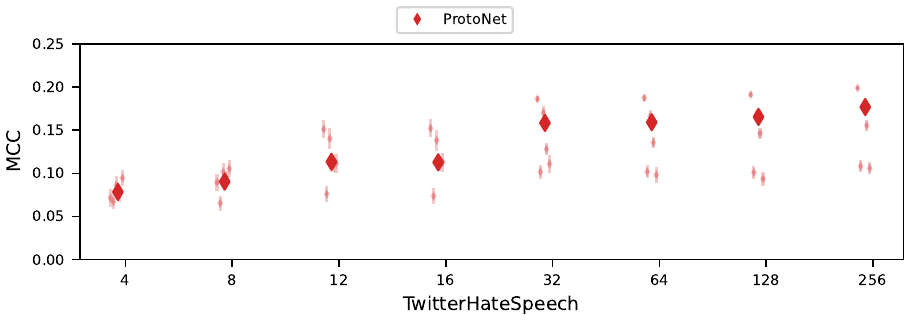}
      \caption{TwitterHateSpeech results using only \textsc{protonet} at much larger values of $k$.}\
      \label{fig:extreme_kshot}
\end{figure*}

To test the capacity of the meta-learners, a limited extension of the TwitterHateSpeech experiment was conducted. Instead of limiting ourselves to $k=16$ examples, we increased to $k=256$. Only \textsc{protonet} was used. Results are depicted graphically in Figure \ref{fig:extreme_kshot} and given numerically in Table \ref{tab:twitter_extreme_kshot}.

We fully expect diminishing returns. Our graph setting implies that the $k$ labelled nodes are already present in the support graph, just with its label masked. Unmasking additional labels should provide little additional information to the model; a good graph learner can already infer the masked labels. In fact, as discussed in Section \ref{sec:transfer}, under heterophily, one might expect the additional labels to provide additional noise for the innocuous class.

Precisely this can be observed in Table \ref{tab:twitter_extreme_kshot}. While the MCC score does increase steadily, it comes at the cost of reduced F1 in the `None' class, and stagnation in the `Racism' class. The only class that sees improvements at very high $k$-shot values, is the homophilic `Sexist' class.

\begin{table*}[t]
    \centering
    \begin{tabular}{ccccc}
        \toprule
        \multirow{2}[2]{*}{\textbf{k}} & \multicolumn{3}{c}{\textbf{F1}} & \multirow{2}[2]{*}{\textbf{MCC}} \\
        \cmidrule(l{4pt}r{4pt}){2-4}
        & \multicolumn{1}{l}{\textbf{Racism}} & \multicolumn{1}{l}{\textbf{Sexism}} & \multicolumn{1}{l}{\textbf{None}} &  \\
        \midrule
        4     & \makecell{\small{0.1949}\\[-5pt] \footnotesize{(0.1880, 0.2019)}} & \makecell{\small{0.2355}\\[-5pt] \footnotesize{(0.2299, 0.2411)}} & \makecell{\small{0.3930}\\[-5pt] \footnotesize{(0.3851, 0.401)}} & \makecell{\small{0.0784}\\[-5pt] \footnotesize{(0.0742, 0.0826)}} \\
        8     & \makecell{\small{0.2157}\\[-5pt] \footnotesize{(0.2092, 0.2222)}} & \makecell{\small{0.2221}\\[-5pt] \footnotesize{(0.2162, 0.2281)}} & \makecell{\small{0.4065}\\[-5pt] \footnotesize{(0.3991, 0.4138)}} & \makecell{\small{0.0904}\\[-5pt] \footnotesize{(0.0862, 0.0946)}} \\
        12    & \makecell{\small{0.2798}\\[-5pt] \footnotesize{(0.2743, 0.2852)}} & \makecell{\small{0.2612}\\[-5pt] \footnotesize{(0.2554, 0.267)}} & \makecell{\small{0.3803}\\[-5pt] \footnotesize{(0.3731, 0.3876)}} & \makecell{\small{0.1134}\\[-5pt] \footnotesize{(0.1092, 0.1177)}} \\
        16    & \makecell{\small{0.2883}\\[-5pt] \footnotesize{(0.2838, 0.2927)}} & \makecell{\small{0.2579}\\[-5pt] \footnotesize{(0.2523, 0.2635)}} & \makecell{\small{0.3740}\\[-5pt] \footnotesize{(0.3664, 0.3816)}} & \makecell{\small{0.1128}\\[-5pt] \footnotesize{(0.1087, 0.117)}} \\
        32    & \makecell{\small{0.3374}\\[-5pt] \footnotesize{(0.3365, 0.3382)}} & \makecell{\small{0.3126}\\[-5pt] \footnotesize{(0.3099, 0.3153)}} & \makecell{\small{0.2947}\\[-5pt] \footnotesize{(0.2902, 0.2993)}} & \makecell{\small{0.1585}\\[-5pt] \footnotesize{(0.1557, 0.1613)}} \\
        64    & \makecell{\small{0.2995}\\[-5pt] \footnotesize{(0.2992, 0.2997)}} & \makecell{\small{0.3092}\\[-5pt] \footnotesize{(0.3068, 0.3115)}} & \makecell{\small{0.2798}\\[-5pt] \footnotesize{(0.2757, 0.2838)}} & \makecell{\small{0.1593}\\[-5pt] \footnotesize{(0.1567, 0.1619)}} \\
        128   & \makecell{\small{0.2971}\\[-5pt] \footnotesize{(0.2970, 0.2973)}} & \makecell{\small{0.3239}\\[-5pt] \footnotesize{(0.3219, 0.3259)}} & \makecell{\small{0.2740}\\[-5pt] \footnotesize{(0.2704, 0.2775)}} & \makecell{\small{0.1655}\\[-5pt] \footnotesize{(0.1632, 0.1677)}} \\
        256   & \makecell{\small{0.3001}\\[-5pt] \footnotesize{(0.2999, 0.3003)}} & \makecell{\small{0.3432}\\[-5pt] \footnotesize{(0.3416, 0.3449)}} & \makecell{\small{0.2713}\\[-5pt] \footnotesize{(0.2680, 0.2746)}} & \makecell{\small{0.1770}\\[-5pt] \footnotesize{(0.1750, 0.1789)}} \\
        \bottomrule
    \end{tabular}
    \caption{TwitterHateSpeech transfer results using only \textsc{protonet} at much larger values of $k$.}
    \label{tab:twitter_extreme_kshot}
\end{table*}

\end{document}